\documentclass[11pt]{article}

\usepackage{acl}

\usepackage{times}
\usepackage{latexsym}

\usepackage[T1]{fontenc}

\usepackage[utf8]{inputenc}

\usepackage{microtype}

\usepackage{inconsolata}

\usepackage{graphicx}
\usepackage{amsmath} 
\usepackage{amssymb} 
\usepackage{booktabs} 
\usepackage{multirow} 
\usepackage{multicol} 
\usepackage{makecell} 
\usepackage{tcolorbox}
\usepackage{subfigure}
\title{CHAI for LLMs: Improving Code-Mixed Translation in Large Language Models through Reinforcement Learning with AI Feedback}

\author{
 \textbf{Wenbo Zhang\textsuperscript{1}},
 \textbf{Aditya Majumdar\textsuperscript{1}},
 \textbf{Asif Ekbal\textsuperscript{2}},
 \textbf{Amulya Yadav\textsuperscript{1}}
\\
\\
 \textsuperscript{1}Pennsylvania State University,
 \textsuperscript{2}Indian Institute of Technology Patna
\\
}

\begin{document}
\maketitle
\begin{abstract}
Large language models (LLMs) show strong performance across many tasks but remain weak at understanding code-mixed (CM) language. Despite this limitation, improving LLMs for CM tasks has received little attention. To address this gap, we propose CHAI, a general-purpose framework for enhancing LLM performance on CM tasks, focusing on CM translation.
CHAI leverages four key ideas. First, we investigate the use of LLMs as annotators to address the scarcity of high-quality CM datasets. Second, we leverage these LLM annotations to generate large-scale preference data and apply reinforcement learning from AI feedback (RLAIF) to improve CM translation. Third, we incorporate LLM-generated domain knowledge as a constitution, enabling iterative response refinement. Fourth, we perform extensive evaluations on real-world datasets and settings.
Results show that CHAI-powered LLMs outperform state-of-the-art open-source models by 68.45\% on average in human-adjudicated win rate on CM translation tasks. This work is a step toward more inclusive open-source code-mixed LLMs. Code: \url{https://tinyurl.com/5fanr47y}.

\end{abstract}

\section{Introduction}

Large language models (LLMs) have excelled at comprehending, generating, and interacting with human language across a wide variety of real-world use cases, e.g., formulating therapeutic dialogue in mental health~\cite{cheng2023now}, 
generating hypotheses in biology~\cite{park2024can}, etc.

Unfortunately, the vast linguistic diversity across the globe still poses significant challenges for LLM-based technologies. In particular, recent studies \cite{winata2026can,yang-chai-2025-codemixbench,mohamed2025lost,zhang2023multilingual,gupta2024codemixeryanahinovel} have shown that the ability of current-day LLMs to understand and generate language is heavily skewed towards monolingual English language queries, with a significant performance degradation reported in prior work \cite{gupta2024code} on tasks involving CM language\footnote{Code-mixing, the fluid alternation between languages within a conversation, is common in multilingual societies.}. These results are highly problematic because they leave a large proportion of the global population - those using CM language as their primary means of communication (with more than 1 billion people in India alone) - at a comparative disadvantage \cite{ramzan2021study}. Ensuring that LLMs benefit these populations requires that future models can understand, reason, and engage with CM languages.

In part, this performance degradation on CM tasks occurs because most current-day LLMs are trained on large corpora of monolingual and/or multilingual text, with comparatively little explicit CM corpora included during the LLM pre-training phase. This lack of inclusion of CM corpora can be attributed to a (relative) lack of availability of large-scale CM datasets on the Internet \cite{magueresse2020low}. 
This presents a significant challenge, as recent model alignment approaches used to improve general LLM performance critically depend on human-annotated preference data, which is extremely scarce for CM tasks.

These challenges motivate us to explore - \emph{Can we develop a general-purpose approach to improve the capability of LLMs in dealing with code-mixed tasks?} To tackle this research question, we propose CHAI (\textbf{C}ode Mixed Understanding via \textbf{H}ybrid \textbf{A}I \textbf{I}nstruction), a novel general-purpose framework for improving the ability of LLMs to handle CM tasks (\emph{we primarily focus on the task of generating CM translations, i.e., translating monolingual text to CM text}). CHAI relies on four novel contributions. First, we investigate the ability of LLMs to provide accurate annotations for CM translation tasks.
We compare LLM annotation results with human annotations, and our results show that LLM labeled preferences (for CM text) are highly correlated with human annotator preferences. 
Second, we leverage this ability of LLMs (to serve as a proxy annotator) to generate preference data for CM translation tasks at scale, which is then used to develop a new code-mixed LLM through model alignment. In particular, we adopt a reinforcement learning from AI feedback (RLAIF) procedure to improve the capability of current-day LLMs to handle CM language. To the best of our knowledge, we are the first to utilize model alignment for the code-mixing scenario. Third, to further improve the capability of LLMs in CM translation tasks, we incorporate LLM-generated domain knowledge and use it as a constitution to iteratively refine the model’s outputs, enabling the LLM to better align its responses with domain-specific knowledge rules. 
Fourth, we conduct rigorous experimental evaluation across various real-world datasets and settings. Our analysis shows that LLMs powered with CHAI outperform conventional state-of-the-art LLMs by 68.45\% on average (in terms of win rate adjudicated by human annotators) on CM translation tasks. We also analyze CHAI's effectiveness on other CM tasks, such as sentiment analysis where it outperforms state-of-the-art baselines by $\sim$37\%. 

\section{Related Work}
\noindent\textbf{LLMs on Code-Mixed Tasks.} A significant body of prior work focuses on evaluating LLMs' ability on CM tasks.~\citet{winata2026can} evaluate LLM abilities in reasoning with CM text, and find that CM perturbations negatively impact the reasoning traces of LLMs, introducing unsupported claims, illogical leaps, and ambiguous facts. Similarly, recent research on benchmarking \cite{yang-chai-2025-codemixbench,mohamed2025lost,zhang-etal-2023-multilingual,gupta2024codemixeryanahinovel} shows that LLMs consistently under-perform on CM datasets across a diverse set of NLP tasks and language pairs. 
Unfortunately, while all these existing studies focus on benchmarking LLMs on CM tasks, none offer solutions for improving performance on such tasks.

\noindent\textbf{Model alignment in machine translation.} A canonical model alignment technique is reinforcement learning from human feedback, or RLHF \cite{ouyang2022traininglanguagemodelsfollow}, which focuses on (i) using human preference data to train a reward model which learns to predict human preferences; and (ii) fine-tuning LLMs using the reward model to align its outputs with user expectations. One line of related work focuses on improving reward modeling for machine translation. For example,~\citet{xu2024advancingtranslationpreferencemodeling} construct reward models by
contrasting deficiencies in machine translation compared to human translation from published books.~\citet{he-etal-2024-improving} utilize the quality estimation (QE) model as their reward model, and show that QE-based feedback training is highly data-efficient. Similarly,~\citet{lai-etal-2024-alarm} introduce hierarchical rewards in RLHF, and show the effectiveness of their approach on machine translation tasks. Unfortunately, prior work focuses solely on monolingual MT tasks. In contrast, we focus on CM machine translation.

\noindent\textbf{RLAIF (Reinforcement Learning from AI Feedback).} Collecting human preference data at scale for RLHF is expensive and time-consuming. Thus, some recent work attempts to replace human feedback with AI (or LLM) feedback, which is then used as preference data to power the conventional RLHF training procedure. ~\citet{bai2022constitutional} first introduced this RLAIF procedure, where an AI labeler identified harmful or harmless outputs to construct a reward model for policy/model optimization. ~\citet{lee2024rlaifvsrlhfscaling} show that RLAIF achieves human-level performance on text summarization and dialogue generation tasks. ~\citet{li2024hrlaifimprovementshelpfulnessharmlessness} propose phased annotations on different prompt categories during the AI preference labeling process, greatly improving the accuracy of AI annotations. To the best of our knowledge, our paper represents the first attempt at adapting RLAIF to improve the ability of LLMs on CM language.

\noindent\textbf{Self-critique in LLMs.} Recent studies explore self-critique mechanisms that enable LLMs to iteratively improve their outputs through self-generated feedback. For example, ~\citet{shinn2023reflexion} and ~\citet{madaan2023self} introduce techniques in which LLMs generate self-feedback and refine their responses without external supervision. Beyond critique to directly improve LLMs' output, some methods leverage synthetic preference signals to further enhance LLMs' performance through the model alignment process. For example,~\citet{dong2024self} and ~\citet{zeng2025aries} present a self-boosting framework that constructs synthetic preference data to train reward models by iteratively alternating between a \emph{self-prompt generator} which produces diverse prompts and a \emph{response improver} which progressively refines model outputs. This process allows LLMs to learn reward models from their own responses, reducing reliance on human preference labels. These methods show that iterative self-feedback and synthetic preference supervision can improve LLM outputs without extensive human annotation. However, they have not been applied to CM scenarios (focus of our work).

\section{CHAI: RLAIF for Code-Mixing}

A key obstacle to scaling RLHF is that the underlying reward model quality depends on high-quality human preference labels, which are costly and time-consuming to collect.
To remedy this issue, RLAIF pipelines \cite{lee2024rlaifvsrlhfscaling} replace human annotators with LLM annotators to generate large-scale preference data for reward model training.

Unfortunately, collecting human annotations for preferences in CM translation is even more challenging because it is harder to find human annotators fluent in two languages. Also, annotation of CM translations is highly subjective (given the informal context in which CM language is used).

\begin{figure*}[ht]
  \centering
  \includegraphics[width=0.9\linewidth]{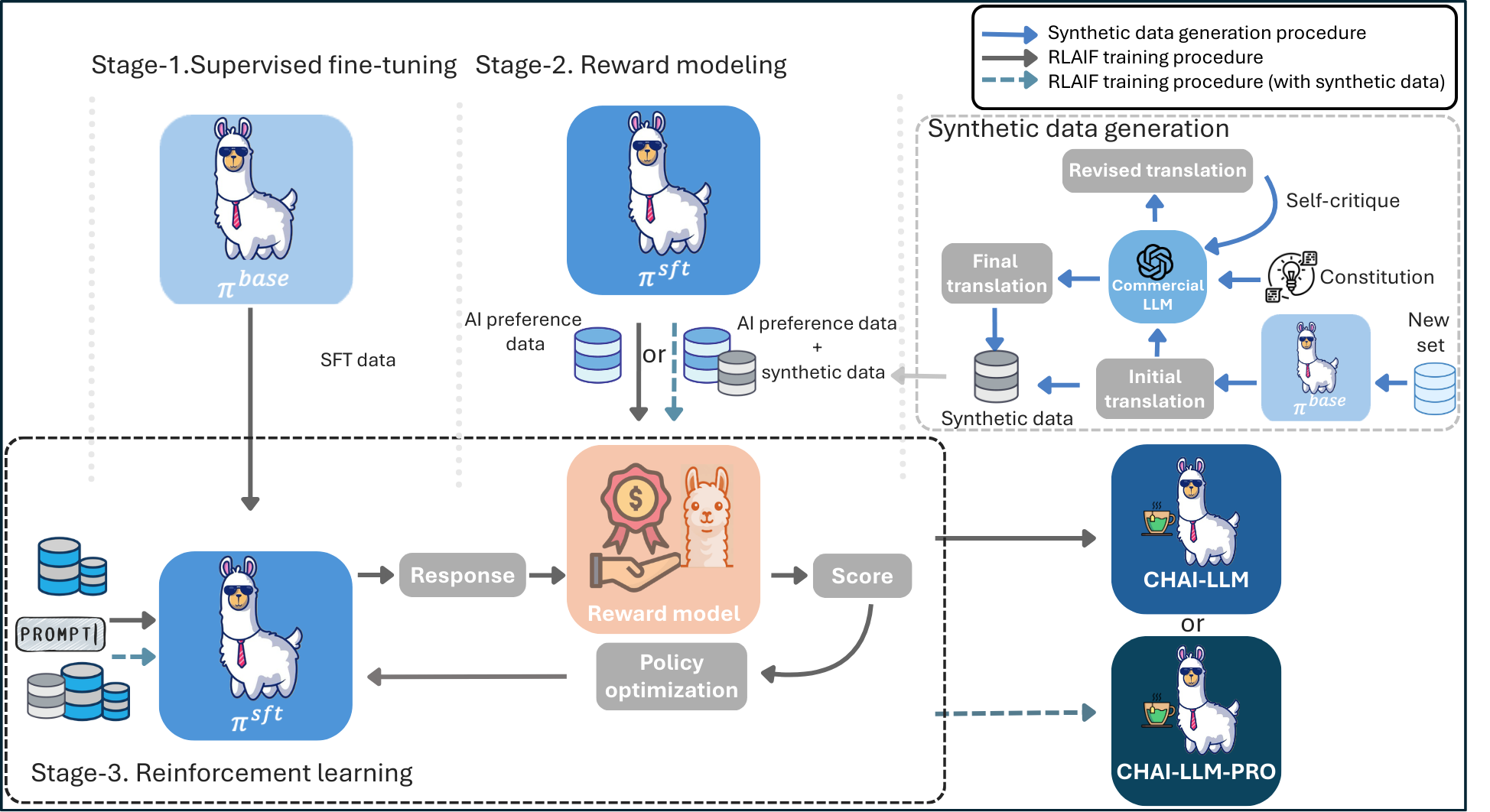}
  \caption {Overall architecture of the RLAIF procedure used in CHAI.}
  \label{fig:RLAIF_procedure}
\end{figure*}

Thus, we propose CHAI, a novel general-purpose RLAIF framework to improve the ability of LLMs to handle CM scenarios. To the best of our knowledge, CHAI is the first to apply RLAIF (or RLHF) for CM scenarios. In particular, CHAI focuses on using RLAIF to improve LLMs' alignment on the task of CM translation using AI-annotated preference labels. Next, we describe CHAI's overall architecture (see Figure~\ref{fig:RLAIF_procedure}).

\noindent \textbf{Base LLM Model.} CHAI starts by using an existing off-the-shelf LLM as a base model (referred to as 
$\pi^{base}$ in Figure \ref{fig:RLAIF_procedure}), which is then further optimized (or aligned) using RLAIF. 

\noindent \textbf{Stage 1: Supervised Fine Tuning of Base Model.} Next, we use the base model and conduct supervised fine-tuning on it using domain-specific data (for CM translation) to adapt the base LLM to the target task (of translating monolingual text into CM text). More formally, given a parallel corpus $\mathcal{D}_{\mathrm{parallel}}=\{(x^{(i)},y^{(i)})\}_{i=1,...,n}$ where $x_{i}$ represents the source (English) sentences, and $y_{i}$ represents the corresponding CM translation, we apply a fixed prompt template $\mathcal{I}_{sft}$ (see Appendix~\ref{tab:prompt_template_for_parallel_corpus}) on this parallel corpus and convert it into a training set $\mathcal{D}_{\mathrm{sft}}=\{(\mathcal{I}_{sft}(x^{(i)}),y^{(i)})\}_{i=1,...,n}$ that can be used to fine-tune our instruction-tuned base model (see Section 3.1 for details). In particular, $\pi^{\mathrm{base}}$ is supervised fine-tuned (SFT) using a next-token prediction objective on this training set $\mathcal{D}_{\mathrm{sft}}$~\cite{radford2019language}. This SFT version of the base model is referred to as $\pi^{sft}$ in Figure 1.

\noindent \textbf{Stage 2: Reward Model Training using AI Feedback.}
In RLAIF, an AI or LLM replaces human annotators to generate preference data. This data trains the reward model, while the remaining pipeline follows RLHF. We now describe how CHAI applies this to CM translation.

\noindent \textbf{2.1 Collecting Preference Data Using LLMs}
We use the $\mathcal{D}_{\mathrm{parallel}}$ corpus (from Stage 1) and convert it into a preference dataset as follows: (i) each source monolingual sentence is paired with two alternative CM translations; (ii) these three sentences are fed into a prompt template $\mathcal{I}_{pref}$ (see Appendix ~\ref{sec:appendix_prompts_for_preference_labeling}) that generates a custom prompt for an LLM annotator asking it to select which of the two provided translations is a better CM translation for the source monolingual sentence. To mitigate positional bias ~\cite{pezeshkpour2023large,li2024hrlaifimprovementshelpfulnessharmlessness} in preference labeling of CM text, we randomly switch the position of the two candidate CM translations before presenting them to the LLM annotator (see Appendix~\ref{sec:appendix_a_positional_bias} for more details on positional bias).

Our final preference dataset contains 15190 distinct prompts (of type $\mathcal{I}_{pref}$) that can be passed to an LLM annotator to get a preference label. To the best of our knowledge, this represents the first-ever attempt at utilizing LLM annotation abilities for annotating tasks related to code-mixing.

\noindent \textbf{2.2 Reward Model Training } This LLM-annotated preference label dataset is then used to train a reward model (RM), which outputs numerical scores for LLM responses provided as input. Intuitively, the trained RM should be such that LLM responses that are closely (or weakly) aligned with AI preferences (expressed in our preference dataset) should receive high (or low) RM scores.

In CHAI, we train our RM as follows: (i) we take our SFT model $\pi^{\mathrm{sft}}$ and change its last neuronal layer from a language modeling head (i.e., output logit of each vocabulary token) into a linear layer which generates a scalar prediction representing the output reward score. (ii) Next, this modified version of $\pi^{\mathrm{sft}}$ is trained on the LLM-annotated preference dataset using the Bradley-Terry model~\cite{bradley1952rank}, which provides a functional form for the probability that for a source (monolingual) sentence $x$, the LLM labeler prefers its chosen CM translation $y_c$ over the rejected translation $y_r$: $P\{i\succ j\}=\frac{e^{r(x, y_c)}}{e^{r(x, y_c)}+e^{r(x, y_r)}}$, 
where $r(x, y_c)$ and $r(x, y_r)$ denote the RM scores for the chosen and rejected CM translations, respectively. This probability is then used in a negative log-likelihood loss: $\mathcal{L}(r)=-\mathbb{E}_{\mathcal{D}_{\mathrm{rm}}}[\log P\{i\succ j\}]$
, where $\mathcal{D}_{\mathrm{rm}}=\{x^{(i)},y_{\mathrm{c}}^{(i)},y_{\mathrm{r}}^{(i)}\}_{i=1}^N$ is the LLM-annotated preference dataset.

\noindent \textbf{Stage 3: Tuning Policy Model with RL. }
Finally, we train a policy model $\pi^{\mathrm{rl}}$ (initialized from $\pi^{\mathrm{sft}}$) to maximize the expected RM score by using general-purpose RL algorithms. More precisely, we optimize the policy model $\pi^{\mathrm{rl}}$ to maximize $r_{total} = r(x,y)-\eta KL(\pi^{\mathrm{rl}}(y|x)||\pi^{\mathrm{sft}}(y|x))$, 
where $r$ refers to the RM score based on a single sample, and the KL divergence term (i) acts as an entropy bonus, preserving generation diversity
~\cite{jaques2019way}; while (ii) ensuring that $\pi^{\mathrm{rl}}$ does not deviate drastically from $\pi^{\mathrm{sft}}$ ~\cite{laidlaw2024preventing,wang2024secrets}. Finally, $\eta$ trades-offs the two objective function terms.

\noindent \textbf{3.1 Architecture Details. }As CHAI's base model, we experiment with Llama-3.1-8B-Instruct \cite{grattafiori2024llama3herdmodels} and Qwen-2.5-7B-Instruct~\cite{qwen2.5} as both are (i) robust multilingual LLMs (supporting English, Hindi, Spanish, etc.); and (ii) have demonstrated strong performance in machine translation tasks \cite{xu2023paradigm}, making them ideal choices for an RLAIF-driven CM translation pipeline.

Given the widespread prevalence of CM language usage in India (in the form of Hinglish, or Hindi+English) \cite{thara2018code}, we focus on datasets for English $\rightarrow$ Hinglish translation in CHAI to power our SFT stage (Stage 1) and our preference data collection stage (Stage 2.1). In particular, we utilize \textbf{MixMT 2022 shared task} \cite{srivastava-singh-2022-overview} and \noindent\textbf{ALL-CS} \cite{tarunesh2021machine} datasets as our parallel corpus $\mathcal{D}_{\mathrm{parallel}}$, which contain $\sim$1800 and 9290 English sentences (respectively), along with multiple Hinglish translations for each sentence.

For each dataset, we first pair each English sentence with each of the available Hinglish translations, and this results in a total of 3873 data points (from the MixMT dataset) + 11317 data points (from the All-CS dataset) = 15190 datapoints inside our parallel corpus  $\mathcal{D}_{\mathrm{parallel}}$.

In Stage 2, CHAI uses GPT-4o \cite{openai2024gpt4technicalreport}  
as an LLM annotator. Each prompt is passed to GPT-4o at three temperature settings (T=0.1, 0.3, 0.5) to get three preference labels, and a majority vote determines the final binary preference label (Y=0 or 1 denotes an LLM preference for the first or second CM translation, respectively).

Finally, in Stage 3, we mainly use proximal policy optimization (PPO)~\cite{schulman2017proximalpolicyoptimizationalgorithms} in CHAI (as our choice of RL algorithm), but we also conduct an ablation with DPO~\cite{rafailov2023direct} in Appendix~\ref{sec:appendix_comparision_dpo}.

\section{Experimental Evaluation}
We evaluate CHAI's ability to improve the performance of our base models for English $\rightarrow$ Hinglish translation. Although CHAI is general enough for other CM language pairs, we focus on this setting because MixMT 2022 and All-CS are the only publicly available datasets providing multiple Hinglish translations per English sentence, which are necessary for generating preference labels in Stage 2 (Figure \ref{fig:RLAIF_procedure}). Collecting such data for other language pairs from human annotators is non-trivial, so we leave this to future work. Nevertheless, we examine the cross-lingual and cross-task transfer ability of our CHAI-powered LLM on additional language pairs (Table \ref{tab:corss_lingual_transfer}) and tasks (Table \ref{tab:performance_sa}).

\noindent \textbf{Baselines. } We compare the performance of CHAI against three baselines: (i) $\pi^{base}$, our base LLM model; (ii) $\pi^{sft-1}$, an SFT version of our base model (as discussed in Stage 1, Figure~\ref{fig:RLAIF_procedure}); (iii) $\pi^{sft-2}$, our second SFT baseline which applies an additional SFT step on LlaMA-3.1-8B or Qwen-2.5-7B (i.e., the non-instruction tuned base version of our base LLM models). This second SFT baseline $\pi^{sft-2}$ is included to isolate any quality degradation from applying SFT to an instruction-tuned model (as has been done in $\pi^{sft-1}$ and CHAI). The training details are in Appendix~\ref{sec:appendix_training_details_sft}.

\noindent \textbf{Evaluation Metrics.} To evaluate CM translation, we utilize six well-studied metrics (see definitions and details in Appendix~\ref{sec:appendix_metirc_explanation}): (i) \textit{BLEU}~\cite{papineni-etal-2002-bleu}; (ii) \textit{chrF}~\cite{popovic2015chrf}; (iii) \textit{chrF++}~\cite{popovic2017chrf++}; (iv) \textit{METEOR}~\cite{banerjee-lavie-2005-meteor}; (v) \textit{COMET}~\cite{rei2020comet}; (vi) \textit{BERTScore}~\cite{zhang2019bertscore}. For sentiment analysis tasks, we use two classic metrics: (i) accuracy; (ii) weighted F1-score.

In addition, we also utilize human and LLM evaluators to calculate the \emph{win rate} for each model, which is the proportion of test cases where evaluators preferred a model’s Hinglish translation over others~\cite{lee2024rlaifvsrlhfscaling}. To compute \emph{win rate} with human evaluators, three human evaluators\footnote{The study was approved by an Institutional Review Board} fluent in both English and Hinglish were recruited. For each source English sentence in the test set, we generated three Hinglish translations from: (i) the CHAI-powered LLM; (ii) the best performing SFT baseline, $\pi^{sft-2}$ (based on performance results in Table~\ref{tab:performance_RLAIF_machine_translation}); and (iii) the base LLM ($\pi^{base}$).
The translations were presented in random order, and evaluators selected their preferred translation; the final label was determined by majority vote (we calculate inter-annotator alignment scores in Appendix \ref{sec:appendix_f}). Similarly, to calculate the win rate with LLM evaluators, we generated three Hinglish translations for each test data point (as described above) and presented them in random order to a 
Gemini-2.5-Flash model~\cite{comanici2025gemini} 
 across three temperature settings (T=0.1, 0.3, 0.5), and aggregated results using a majority vote.

\noindent \textbf{Evaluation Datasets.} 
 Machine translation experiments use the test sets of the widely used PHINC~\cite{srivastava-singh-2020-phinc} and MT-AUG~\cite{dhar-etal-2018-enabling} datasets. Cross-lingual transfer experiments rely on corpora from~\cite{gupta2024multilingual}. Sentiment analysis is evaluated on the SentMix-3L dataset~\cite{raihan2023sentmix} and the SemEval-2020 Task 9 test set~\cite{patwa2020semeval2020task9overview}. Dataset details are in Table~\ref{tab:info_test_set}.

We present results in four stages. First, we evaluate the ability of LLM annotators to mimic human preferences in CM translation. Second, we compare CHAI-powered LLMs with state-of-the-art baselines on CM translation quality. Third, we analyze transfer learning by evaluating performance on Hinglish sentiment analysis and cross-lingual translation. 
Finally, we study performance scaling with synthetic data generated via LLM self-critique.

\begin{table}[t]
  \centering
  \scalebox{0.75}{
  \begin{tabular}{ccc}
    \bottomrule
    Strategy &Prompt & \makecell{Alignment\\score} \\
    \bottomrule
    & Basic 0-shot & 66.70\%\\
    Basic & Basic 1-shot & 61.36\%  \\
    & Basic 2-shot& 56.14\%  \\
    \hline
        & \textbf{Basic + rule 0-shot} &  \textbf{74.43\%}  \\
    \makecell{Rule-augmented} & Basic + rule 1-shot &  64.20\% \\
    & Basic + rule 2-shot &  57.39\% \\
    \hline
    & Basic + CoT 0-shot & 61.48\%\\
   \makecell{Chain-of-thought} & Basic + rule + CoT 0-shot &  65.23\%  \\
    + & Basic + rule + CoT 1-shot &  64.55\%  \\
    Rules & Basic + rule + CoT 2-shot &  66.93\%  \\

  \bottomrule
\end{tabular}
 }
\caption{Alignment between human \& LLM annotators.
}
\label{tab:performance_llm_labelling_strategy}
\end{table}

\noindent \textbf{LLM Annotator Alignment. } To generate preference labels in Stage 2 (Figure \ref{fig:RLAIF_procedure}), we tested various combinations of three prompting strategies (basic prompts \ref{tab:prompt_preference_annotation_zero_shot_basic}, rule-augmented prompts \ref{tab:prompt_preference_annotation_zero_shot_detailed}, and chain-of-thought prompts \ref{tab:prompt_preference_annotation_cot_basic}) to generate \emph{LLM-annotated preference labels} and compared them against \emph{human-generated preference labels} (three human annotators were also used to provide preference labels on training data points). Table \ref{tab:performance_llm_labelling_strategy} shows alignment scores (fraction of cases where LLM annotations match human annotations) for different prompting strategies. The results show that 0-shot prompting with explicit preference rules (see Table~\ref{tab:prompt_llm_as_judge} for examples of rules) performs best, achieving the highest alignment score of 74.43\%, outperforming other methods by 19.20\% on average. Rule-augmented prompting generally improves LLM–human alignment by 1.28\% on average for CM tasks. Surprisingly, chain-of-thought (CoT) and k-shot prompting do not improve alignment, possibly due to the inconsistent grammatical structure of CM text. We therefore adopt the best-performing prompting strategy from Table~\ref{tab:performance_llm_labelling_strategy}.


\begin{table*}[!ht]
  \centering
  \scalebox{0.70}{
  \begin{tabular}{ccccccccccccc}
    \toprule
     & \multicolumn{6}{c}{MT-AUG} & \multicolumn{6}{c}{PHINC}\\

    \cmidrule(lr){2-7}
    \cmidrule(lr){8-13}
     & BLEU & ChrF & CHrF++ & METEOR & COMET & BERTscore
     & BLEU  & ChrF & CHrF++ & METEOR & COMET & BERTscore\\
    \bottomrule
    
    $\pi^{\mathrm{base}}$ & 1.99  & 25.97 & 22.45 & 0.22 & 0.66 & 0.86 &
    10.59 & 34.70 &  30.98 & 0.24 & 0.64 & 0.88 \\
    \hline
    $\pi^{\mathrm{sft-1}}$ & 2.10 & 23.56 & 21.69 & 0.21 & 0.67 & 0.87 &
    17.13 & 35.02 & 31.45 & 0.27 &  0.65 & 0.88 \\
     \hline
    $\pi^{\mathrm{sft-2}}$ &2.27  & 24.03 & 22.23  & 0.22 & 0.67 & 0.87 & 
    17.44  & 35.69 & 32.20 & 0.28 & 0.65 & 0.88 \\
     \hline
    CHAI-LLM & \textbf{3.66}  & \textbf{34.64}  &  \textbf{30.02} & \textbf{0.28}& \textbf{0.78} &\textbf{0.88}& 
    \textbf{18.11}  & \textbf{40.08} & \textbf{35.47} & \textbf{0.33} & \textbf{0.75} & \textbf{0.89} \\

  \bottomrule
\end{tabular}
 }
\caption{Measuring CHAI’s ability in improving code mixed translation ability.}
\label{tab:performance_RLAIF_machine_translation}
\end{table*}

\begin{figure*}[h!]
\centering

\subfigure[Evaluation on MT-AUG test set]
{
    \centering
    \includegraphics[width=0.48\textwidth]{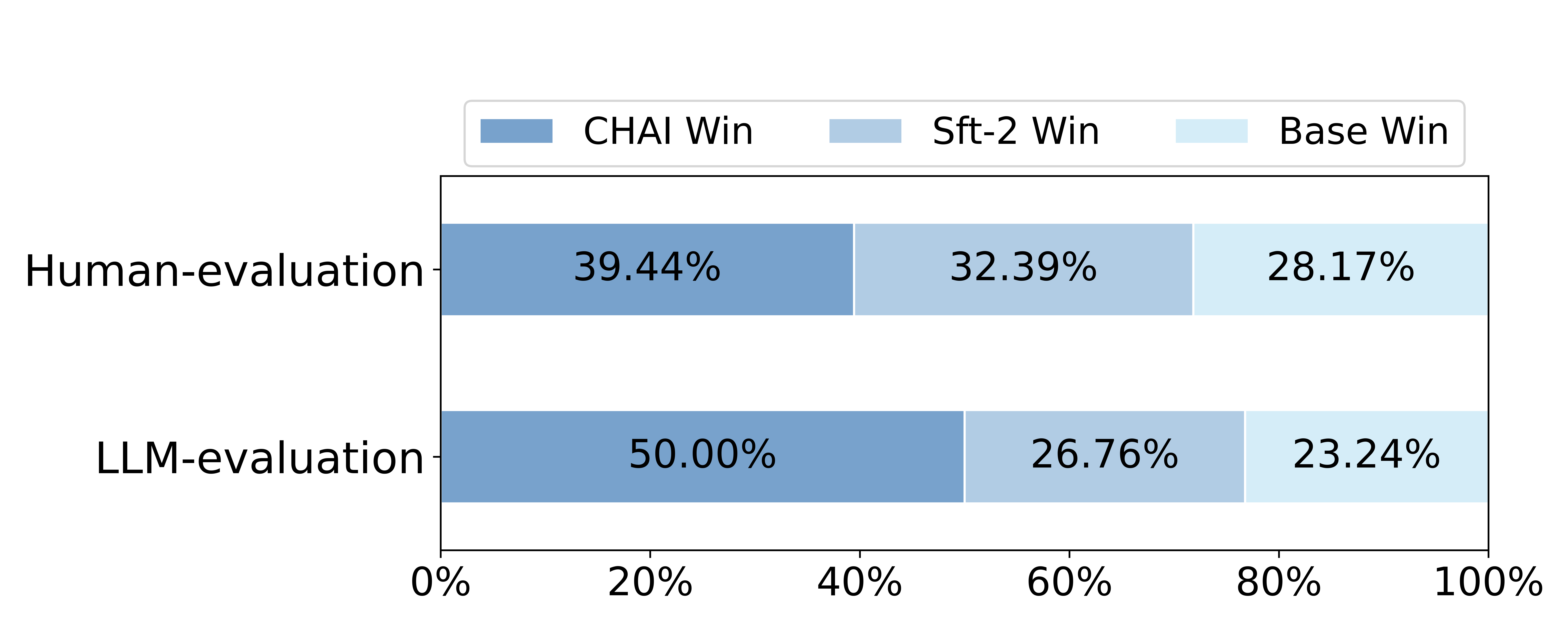}
    \label{fig:win_rate_3group_mtaug}
}
\hfill
\subfigure[Evaluation on PHINC test set]
{
    \centering
    \includegraphics[width=0.48\textwidth]{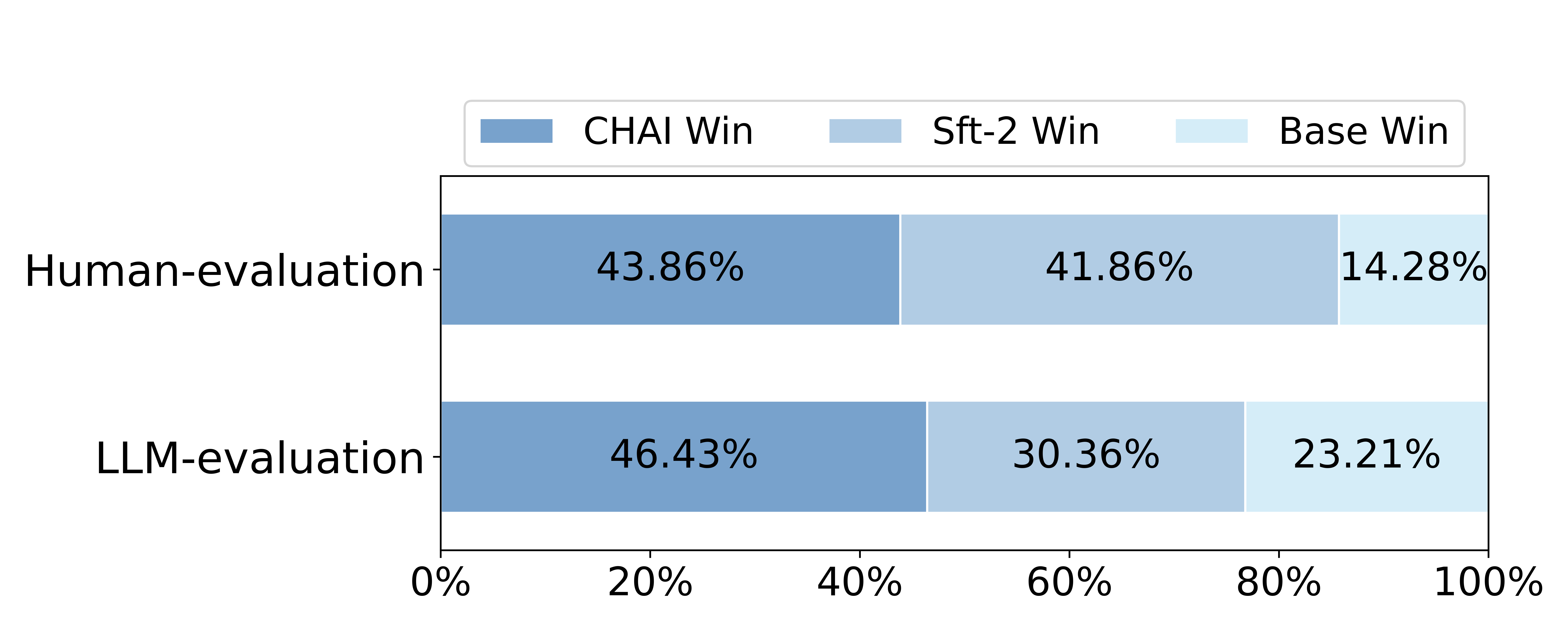}
    \label{fig:win_rate_3group_phinc}
}

\caption{Corresponding win rate to measure CHAI’s ability in improving code mixed translation ability.}
\label{fig:performance_RLAIF_machine_translation_winrate_v3}

\end{figure*}

\begin{table*}[!ht]
  \centering
  \scalebox{0.70}{
  \begin{tabular}{ccccccccccccc}
    \toprule
     & \multicolumn{6}{c}{\makecell{English$\rightarrow$ Bengali+English}} & \multicolumn{6}{c}{\makecell{English$\rightarrow$ Spanish+English}}\\

    \cmidrule(lr){2-7}
    \cmidrule(lr){8-13}
     & BLEU & ChrF & CHrF++ & METEOR & COMET &BERTscore &BLEU  & ChrF & CHrF++ & METEOR & COMET &BERTscore\\
    \bottomrule
    
    $\pi^{\mathrm{base}}$ & 1.46  & 12.75 & 11.42 & 0.09 & 0.59 & 0.84  & 4.12 & 32.72 &  30.56 & 0.38 & 0.65 & 0.89\\
    \hline
    $\pi^{\mathrm{sft-1}}$ & 3.10 & 17.79 & 13.84 & 0.16 & 0.57 & 0.85 & 3.60 & 29.50 & 25.16 & 0.14 &  0.75 & 0.90 \\
     \hline
    $\pi^{\mathrm{sft-2}}$ &3.79  & 18.50 & 16.27  & 0.24 & 0.60 & 0.86 & 4.47  & 31.27 & 27.34 & 0.19 & 0.64 & 0.88 \\
     \hline
    CHAI-LLM & \textbf{4.09}  & \textbf{19.94}  &  \textbf{17.48} & \textbf{0.26}& \textbf{0.66} &\textbf{0.87}& 
    \textbf{5.23}  & \textbf{35.56} & \textbf{33.09} & \textbf{0.45} & \textbf{0.79} & \textbf{0.92}\\

  \bottomrule
\end{tabular}
 }
\caption{Cross-lingual transfer result based on different code-mixed language pairs.}
\label{tab:corss_lingual_transfer}
\end{table*}

\begin{table}[t]
  \centering
  \scalebox{0.75}{
  \begin{tabular}{p{2.0cm}p{1.05cm}p{1.2cm}p{1.2cm}}
    \bottomrule
    Dataset & LLM & Accuracy & F1\_score \\
    \bottomrule
    \multirow{4}{*}{SemEval-2020}&$\pi^{\mathrm{base}}$ & 35.40\% &22.65\%\\
    & $\pi^{\mathrm{sft-1}}$ & 35.97\% & 23.55\%\\
    & $\pi^{\mathrm{sft-2}}$ & 35.60\% & 22.74\%\\
    & CHAI & \textbf{36.77\%} & \textbf{25.04\%}\\
    \hline
    \multirow{4}{*}{SentMix-3L}&$\pi^{\mathrm{base}}$ & 44.39\% & 32.96\%\\ 
    & $\pi^{\mathrm{sft-1}}$ & 46.17\% & 33.91\%\\
    & $\pi^{\mathrm{sft-2}}$ & 45.17\% & 33.19\%\\
    & CHAI & \textbf{55.21\%} & \textbf{46.38\%}\\

  \bottomrule
\end{tabular}
}
\caption{Performance of CHAI on sentiment analysis.}
\label{tab:performance_sa}
\end{table}

\noindent \textbf{Impact of CHAI on Translation Quality.} Having identified the best prompting strategy and temperature setting (see Appendix~\ref{sec:Tuning_LLM_Temperature}), we now train a CHAI-powered LLM with these optimal hyperparameters to evaluate its effectiveness for CM translation. Table \ref{tab:performance_RLAIF_machine_translation} 
compares six evaluation metrics achieved by CHAI-LLM (against three baselines) on MT-AUG and PHINC test sets. CHAI-LLM consistently outperforms all baselines on all metrics, e.g., CHAI achieves 77.47\% (and 24.44\%) improvement in BLEU (and CHrF++) over the base model, respectively. Among SFT baselines, $\pi^{sft-2}$ does marginally better, which illustrates that performing SFT on an instruction-tuned model led to some performance drop see performance analysis in Appendix~\ref{sec:appendix_sft_performance_degradation}).
Nevertheless, CHAI achieves 32.58\% (and 28.23\%) improvement in BLEU (and CHrF++) over $\pi^{sft-2}$, the best-performing SFT baseline model, respectively. Similar results using Qwen 2.5 as the base model are in Appendix~\ref{sec:appendix_qwen}).

Figure~\ref{fig:performance_RLAIF_machine_translation_winrate_v3} compares LLM- and human-evaluated win rates of CHAI-LLM against three baselines on the same test sets. CHAI-LLM achieves an average win rate of 48.22\% (and 41.65\%) against baseline models, as adjudged by LLM evaluators (and human evaluators), respectively. On both datasets, CHAI-LLM was preferred by human annotators (as compared to $\pi^{\mathrm{sft-2}}$, which achieved lower average win rates of 28.56\% and 37.13\%). Overall, CHAI-LLM outperformed state-of-the-art open-source models by 68.45\% (averaged across performance gap over $\pi^{\mathrm{sft-2}}$ and $\pi^{\mathrm{base}}$) in human win rate. Notably, these results use a three-group comparison, and a win rate above 33.33\% indicates competitive performance relative to the others.

Thus, results in Table \ref{tab:performance_RLAIF_machine_translation} and Figure~\ref{fig:performance_RLAIF_machine_translation_winrate_v3} establish that CHAI significantly improves LLM performance on CM translation. Our comparison results are statistically significant, see Appendix~\ref{sec:appendix_stats}) for details. Further, Table~\ref{tab:sampels_machine_translation_after_RLAIF} compares translations from baseline models and CHAI-LLM on two representative examples. As shown in Table~\ref{tab:sampels_machine_translation_after_RLAIF}, CHAI-LLM produces more natural CM translations aligned with phrasing preferred by Hinglish speakers.

To further substantiate our argument, we present three additional code-mixing evaluation metrics for the code-mixed machine translation task in Figure~\ref{tab:performance_RLAIF_machine_translation}: (i) Code-Mixing Index (CMI)~\cite{srivastava2021challenges}; (ii) Multilingual Index (M-index)~\cite{barnett2000lides}; and (iii) I-Index~\cite{guzman2017metrics}. Due to lack of space, these results can be found in Appendix \ref{sec:appendix_additional_code_mixed_metrics}.

\begin{table*}[!ht]
  \centering
  \scalebox{0.75}{
  \begin{tabular}{p{2cm}|p{4cm}|p{8.7cm}}
    \toprule
     & Results & English -> Hinglish \\
    \toprule

     & Input (English) & Failed to load remote file. \\
     Sample-1 & $\pi^{sft-1}$ output (Hinglish) & Durg file load karne mein vifal.  \\
     & CHAI output  (Hinglish) & Remote file load karne mein fail ho gaya hai.  \\
     \cline{2-3}
     & Comments & $\pi^{sft-1}$ uses “Durg” which is not a correct word, making the translation unnatural and harder to understand in everyday Hinglish usage. Instead, CHAI uses the correct form of the sentence with correct grammar as well.\\
     
     \hline
     & Input (English) & Action is the foundational key to all success. \\
     Sample-2 & $\pi^{sft-2}$ output (Hinglish) & Kisi success ka rahna hain.  \\
     & CHAI output  (Hinglish) & Action toh sab success ke liye foundational key hai.  \\
     \cline{2-3}
     & Comments & $\pi^{sft-2}$ omits the relevance of "action" and mistranslates the sentence to  "to remain some success". This is grammatically incorrect and misses the core message. CHAI accurately captures the true meaning of the sentence.\\
     
  \bottomrule
\end{tabular}
}
\caption{Comparing the translations generated from different baselines and the
CHAI-powered LLM.}
\label{tab:sampels_machine_translation_after_RLAIF}
\end{table*}

\noindent \textbf{Cross-lingual Transferability. } Prior research shows that models fine-tuned for machine translation exhibit strong cross-lingual transferability~\cite{lample2019cross}. Motivated by this prior work, we examine if translation preferences learned during CHAI's RLAIF procedure enhance cross-lingual transfer. We evaluate three directions: (i) English $\rightarrow$ English + Bengali (En+Be);(ii) English $\rightarrow$ English + Spanish (En+Es); and (iii) English $\rightarrow$ English + French (En+Fr).

Evaluation results for the first two translation directions are presented in Table~\ref{tab:corss_lingual_transfer}, while results for En→CM (Fr–En) are shown in Appendix~\ref{sec:appendix_additional_Cross_lingual_Transfer}. Across six auto metrics, CHAI-LLM shows strong and consistent performance. For BLEU, CHAI-LLM achieves the best performance in En→CM (Be–En) with a score of 4.09 and En→CM (Es–En) with 5.23, outperforming all baselines. In terms of COMET, CHAI-LLM performs best, scoring 0.66 and 0.79 respectively. Results on chrF, chrF++, METEOR, and BERTScore follow similar patterns.

These results indicate that RLAIF improves CHAI-LLM’s cross-lingual transfer across multiple CM languages, consistent with prior findings on MT-based pretraining for multilingual generalization~\cite{lample2019cross}.

\noindent \textbf{Cross-Task Ability. } 
We explore if using RLAIF for CM translation improves an LLM's general ability to handle additional CM tasks. Table~\ref{tab:performance_sa} compares the accuracy and F1 of our CHAI-powered LLM and three baselines on two CM sentiment analysis datasets with Hinglish input and ternary sentiment labels (positive, neutral, negative).

CHAI outperforms all baselines on SemEval-2020 with 25.04\% F1, improving by 2.30\% F1 over the best baseline. On SentMix-3L, it achieves 46.38\% F1, with gains of 12.47\% F1 over $\pi^{sft-1}$. Results on accuracy also show the same trend on both sets. This table indicates that using RLAIF improves LLMs' ability to handle other CM tasks.

\section{Improving RLAIF via Self-Critique}
Our earlier results show that CHAI, our RLAIF framework, significantly improves machine translation and benefits downstream tasks such as sentiment analysis. However, a key bottleneck in applying RLAIF to CM translation is the scarcity of CM datasets that contain multiple candidate translations per source English sentence, which are needed to generate preference annotations for reward model (RM) training. In our experiments, CHAI relies on the only two publicly available datasets - MixMT 2022~\cite{srivastava-singh-2022-overview} and ALL-CS~\cite{tarunesh2021machine} - that provide multiple CM translations for each source sentence. The limited availability of such datasets constrains further improvements for English $\rightarrow$ Hinglish and other language pairs. To address this, we introduce a self-critique method to efficiently generate additional synthetic preference data, improving both the RM and the RL-tuned LLM.

Inspired by Constitutional AI~\cite{bai2022constitutional}, we propose a self-critique procedure to generate synthetic preference data. The original Constitutional AI framework derived supervision from explicit rules/principles (called a \emph{constitution}) to guide SFT data generation and preference annotation. In contrast, we use a constitution-driven approach to generate synthetic preference pairs for RM training and RL-based translation tuning, addressing the data bottleneck in CM scenarios.

To construct a \emph{constitution} tailored for CM translation, we prompt GPT-5-mini using a template (Table~\ref{tab:prompt_for_ai_institution}) to generate first-principle rules that high-quality CM translations should satisfy. Ideally, such principles would be curated by expert linguists. However, obtaining expert annotations is costly and time-consuming, which partly explains the scarcity of preference data in CM settings. Instead, we adopt a low-cost alternative by generating the constitution with an LLM. While these principles may not be optimal, this low-cost approach enables rapid creation of candidate guidelines without expert supervision. We then evaluate how well these principles guide synthetic preference data generation for CM translation. The resulting principles form the constitution (see Appendix~\ref{sec:appendix_Selected Principles}) that governs our proposed critique–revision process.

With the constitution defined, our synthetic preference generation pipeline proceeds as follows. We first sample English source sentences from the HinGE dataset~\cite{srivastava2021hinge}. For each sentence, an initial CM translation is generated using Llama3.1-8B-Instruct, serving as the baseline response. A stronger commercial LLM (GPT-5-mini, which balances cost with reasoning performance) then critiques the translation based on one constitution principle (selected by GPT-5-mini according to the source sentence and the current translation). Then, GPT-5-mini revises the translation using its self-generated critique.

This critique–revision process is applied iteratively. At each step, GPT-5-mini selects a constitutional principle based on the source sentence and current translation, critiques the translation, and revises it accordingly. After N=3 steps of critique \& revision, the translation is progressively improved according to CM translation principles.

After refinement, we construct synthetic preference pairs consisting of the English source sentence, the initial LLM translation, and the final revised translation. Assuming the revision yields a better CM translation (see Appendix~\ref{sec:appendix_translation_quality_on_synthetic_data} for evidence), we treat it as the preferred response. These pairs are combined with the original CHAI preference dataset to train the reward model and RL-tuned translation model under the same RLAIF setup, and we then evaluate their impact on CM translation quality on MT-AUG and PHINC.

\begin{figure}[t]
\centering

\subfigure[Evaluation on MT-AUG]
{
    \centering
    \includegraphics[width=0.46\columnwidth]{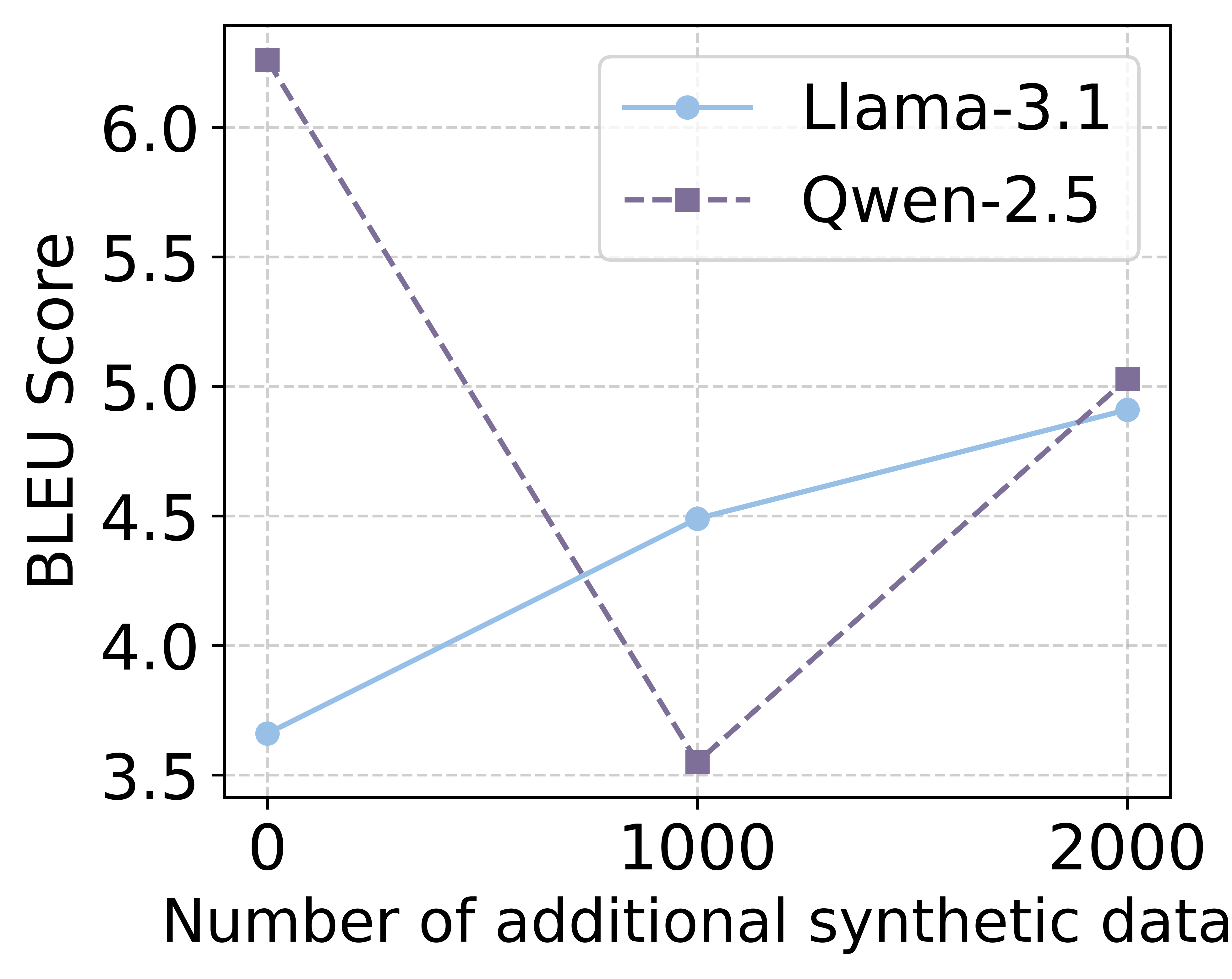}
    \label{fig:additional_synthetic_data_mtaug_bleu_plot}
}
 \hfill
\subfigure[Evaluation on PHINC]
{
    \centering
    \includegraphics[width=0.46\columnwidth]{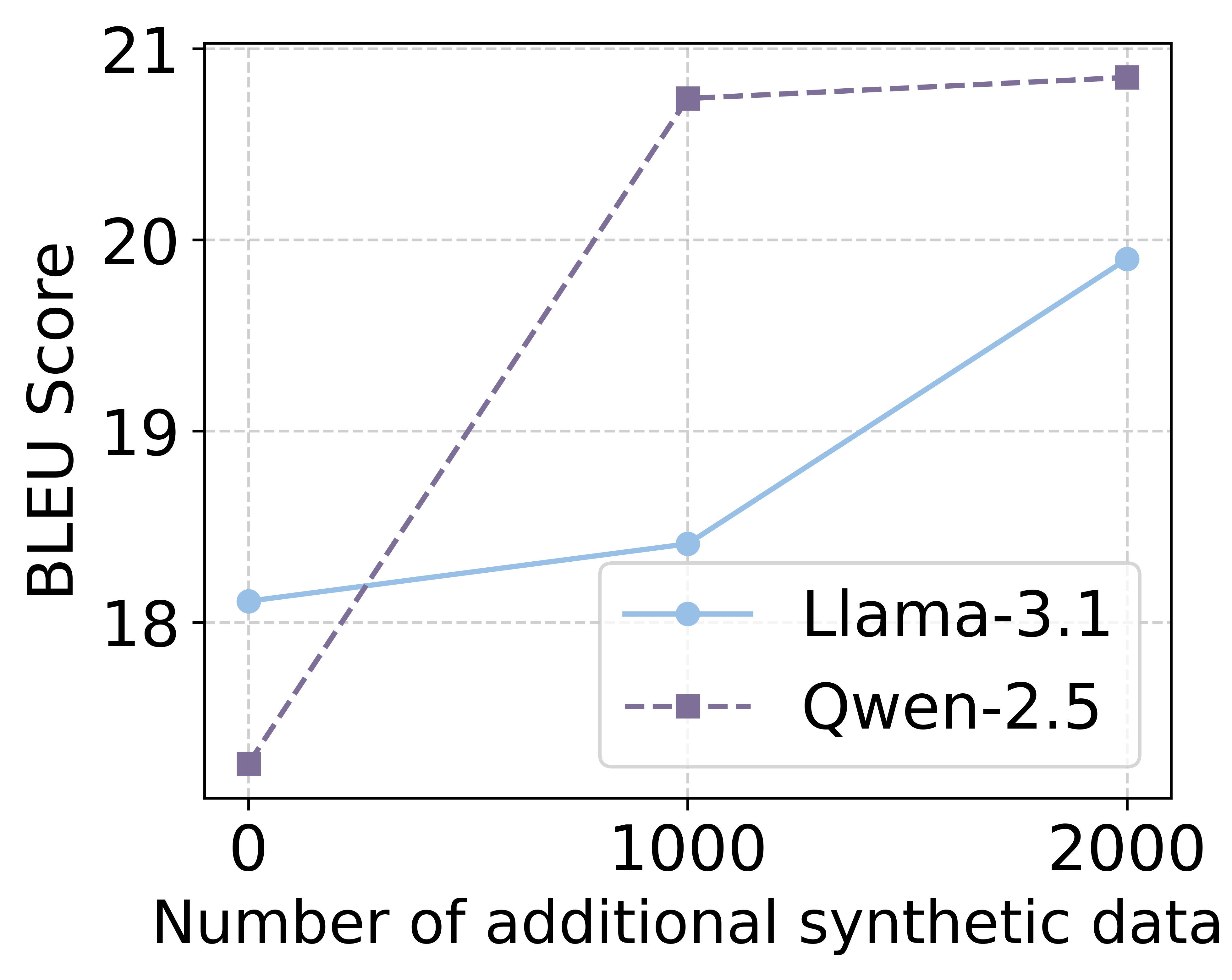}
    \label{fig:additional_synthetic_data_phinc_bleu_plot}
}

\caption{BLEU of CHAI-LLM trained on the original dataset with vs. without additional synthetic samples.}
\label{fig:performance_RLAIF_synthetic_data_bleu}

\end{figure}

Figure~\ref{fig:performance_RLAIF_synthetic_data_bleu} shows the relationship between the amount of synthetic preference data and BLEU score (results for other metrics are in Appendix~\ref{sec:appendix_additional_evlaution_syntheric_data}). Overall, the LLaMA base model show consistent improvements with increasing synthetic preference data for RM training, while this trend is not observed for the Qwen model.

\begin{figure}[t]
  \centering
  \includegraphics[width=0.99 \linewidth]{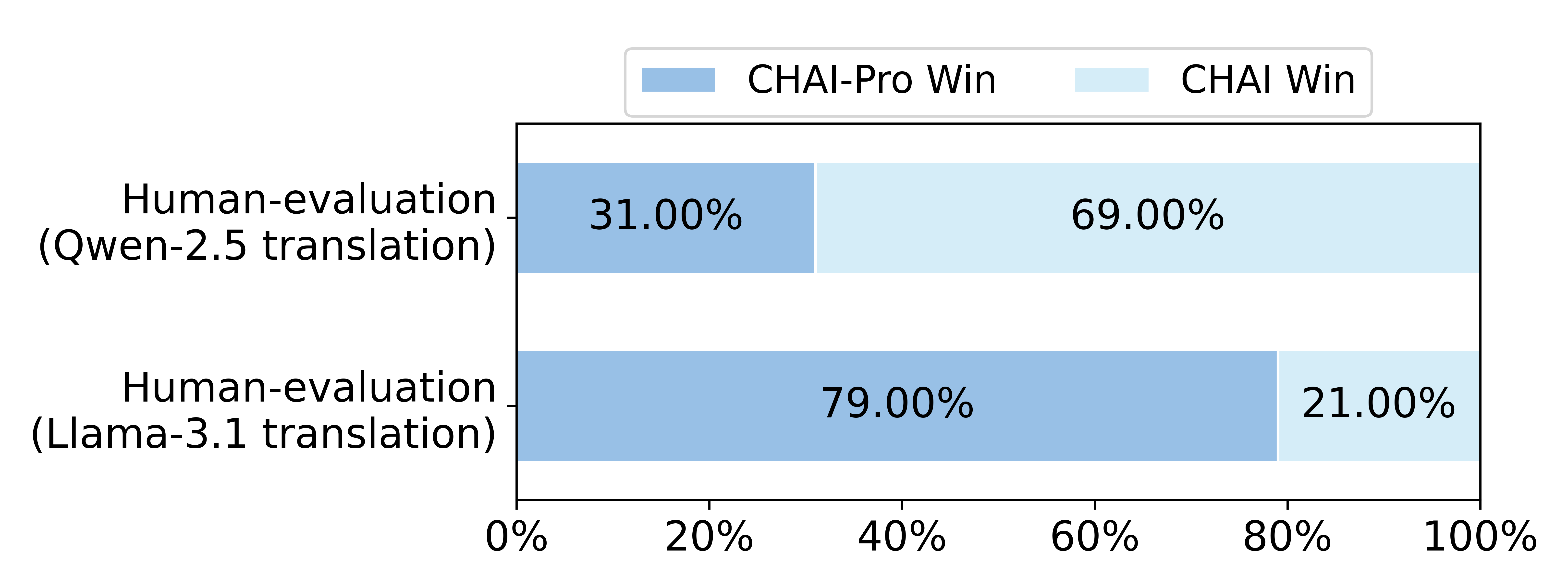}
  \caption {Human win rate comparison of CHAI-LLM-Pro (with 2,000 synthetic samples) VS CHAI-LLM.}
  \label{fig:performance_RLAIF_synthetic_data_winrate}
\end{figure}

We also randomly sample 100 data points for an exploratory human evaluation comparing CHAI-LLM (no synthetic data) with CHAI-LLM-Pro (RM trained with 2,000 synthetic preference pairs). Figure~\ref{fig:performance_RLAIF_synthetic_data_winrate} shows similar results: the LLaMA base model benefits more from synthetic data (79\% win rate) than Qwen base model (31\%). The observed discrepancy suggests that synthetic preference data effectiveness is model-dependent, consistent with recent findings on varying LLM sensitivity to alignment signals~\cite{sharma2024critical}.


\section{Limitations}
Due to the non-trivial effort involved in gathering annotations from professional crowd (human) annotators across different language pairs, this study focuses on a single language pair (Hindi and English) and leave the exploration of other language pairs for future work. This naturally limits our evaluation somewhat. 
Additionally, the study focuses on implementing CHAI on two open-source LLMs (Llama-3.1 and Qwen-2.5) and corresponding versions are less than 10 billion parameters. Conducting experiments with larger base models is highly challenging in an academic research setting due to computational constraints. Therefore, we focused on evaluating one of the most powerful open-source base models available.
Next, the study mainly focuses on a single NLP task during the alignment training procedure: machine translation. Cross-lingual experiments (in Table~\ref{tab:corss_lingual_transfer}) and the sentiment analaysis task (in Table~\ref{tab:performance_sa}) are only included in the evaluation part. 
In future work, we aim to experiment with other directionalities of translation and more general NLP tasks such as CM summarization, word-level language identification, etc. 
Finally, while we recognize that there are other important dimensions for evaluating translation quality such as the presence/absence of bias, helpfulness/harmfulness of translations, etc., this study evaluates performance solely based on translation accuracy. We leave the exploration of these other evaluation dimensions for future work.

\section{Ethical Considerations}
The problem studied in this paper - development of LLMs for code mixed translation - presents several ethical challenges that need to be discussed and contemplated. First, it is important that such CM LLMs output fair and unbiased translation outputs. In particular, it is necessary to be vigilant about situations in which biases in CM training data lead to biased or skewed translations that may end up reinforcing problematic social norms, or misrepresenting cultural nuances. Additionally, preserving the intent and sentiment of speakers is essential, particularly in settings where such CM translations are used to interact with CM speakers. 

Perhaps most importantly, the ethics of circumventing human feedback with AI feedback (as is the norm in RLAIF procedures) needs to be discussed carefully. On one hand, as the results of this paper show, leveraging AI feedback in RLAIF procedures will speed up the developmennt of inclusive CM LLMs which will help bridge the digital divide, by making the benefits of LLMs available to lots of CM speakers from places like South Asia. On the other hand, utilizing AI feedback (in RLAIF) might mean fewer opportunities for human crowd workers (a majority of whom live in South Asia) to provide annotations and receive renumeration in return. Thus, the ethics of leveraging LLMs as annotators deserves serious discussion (especially with regards to the associated negative impacts on the livelihoods of human crowd annotators).

\bibliography{custom}

\appendix

\section{Appendix}
\label{sec:appendix}
\renewcommand{\thetable}{A\arabic{table}}
\renewcommand{\thefigure}{A\arabic{figure}}
\setcounter{table}{0} 
\setcounter{figure}{0}

\subsection{Prompt Template to Create Parallel Corpus}
See template at Table~\ref{tab:prompt_template_for_parallel_corpus}.

\label{sec:appendix_c}
\begin{table}[!ht]
    \centering
    \small
    \begin{tabular}{p{7cm}}
    \toprule

    \textbf{Prompt template $\mathcal{I}$:} \\
    Translate this from \{Source\} to \{Target\}:\\
    \lbrack Source\rbrack: \{x\} \\
    \lbrack Target\rbrack: \{y\} \\

    \bottomrule
    \end{tabular}
    \caption{Prompt template to create parallel corpus, where ’Source’ and ’Target’ represent the names of the source language and the target language,respectively.}
    \label{tab:prompt_template_for_parallel_corpus}
\end{table}

\subsection{Positional Bias in Code-mixed Texts}
\label{sec:appendix_a_positional_bias}

We use the same test set (previously used for Section~\ref{sec:evaluation_set}) to evaluate the positional bias problem in annotating code-mixed texts. For each example in the test set, we ask different LLM labelers to generate preference labels for a pair of candidates through the basic prompt in \ref{tab:prompt_preference_annotation_zero_shot_basic}. Then the candidate order presented in the prompt is swapped, and the same LLMs are requested to generate preference labels again. If an LLM favors the same opinion on both the original and reversed order of candidates in the prompt, we consider it to be biased. 

In this section, we measure position bias by computing the alignment score between the LLM annotated results and human preference labels. From Table~\ref{tab:problem_positional_bias}, we see that both LLM labelers (GPT-4o and Gemini) shows different alignment scores on same preference labeling task. This observation indicates the positional bias of LLM labelers also exists through the preference annotation task on code-mixed texts.

\begin{table}[!ht]
  \centering
  \begin{tabular}{ccc}
    \bottomrule
    LLM labeler & Alignment score \\
    \bottomrule
    GPT-4o (default order) & 59.7\% \\
    GPT-4o (switched order) & 54.3\% \\
    \hline
    Gemini (default order) & 59.0\% \\
    Gemini (switched order) & 55.2\% \\

  \bottomrule
\end{tabular}
\caption{Performance of LLM labelers with different positional orders.}
\label{tab:problem_positional_bias}
\end{table}

\subsection{Details of Evaluation Set for Alignment Score Calculation}
\label{sec:evaluation_set}
We downsampled from the training set $\mathcal{D}_{\mathrm{rm}}$ and create an evaluation set containing 1000 data points. Each data point contains one English sentence and two corresponding code-mixed Hinglish translations. Each sample is assessed by three independent human annotators. The human preference labels are obtained through the majority voting of three human annotators' results.

\subsection{Tuning LLM Temperature}
\label{sec:Tuning_LLM_Temperature}
\begin{figure}[t]
  \includegraphics[width=0.97\linewidth]{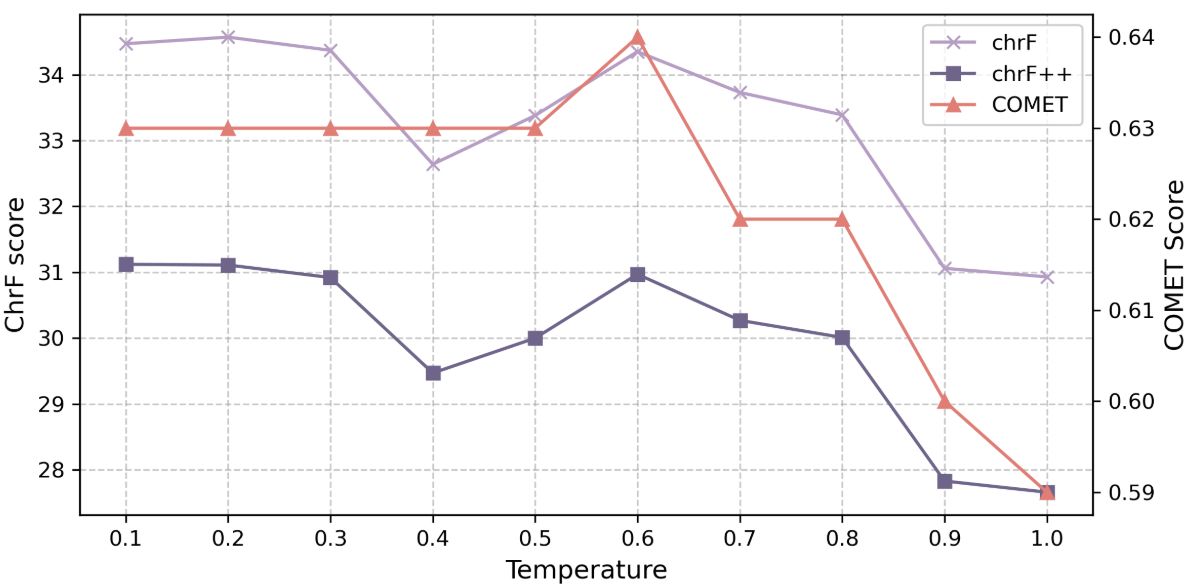}
  \caption {Relationship between the temperature and the quality of code-mixed machine translation.}
  \label{fig:temperature_vs_translation_quality}
\end{figure}

In Figure \ref{fig:temperature_vs_translation_quality}, we compare the variation in code-mixed translation quality (as measured by chrF, chrF++, and COMET on Y-axes) with increasing values of temperature for the CHAI-powered LLM (X-axis). This figure shows that all three metrics are optimized at T=0.6. Thus, we fix the temperature of the CHAI-powered LLM to T=0.6 in all future experiments. 

\subsection{Statistical Information about Test Sets}
\label{sec:appendix_statistical_info}
See details of test sets in Table~\ref{tab:info_test_set}.

\begin{table*}[!ht]
  \centering
  \begin{tabular}{p{2.0cm}p{3.2cm}p{1.8cm}p{1.4cm}p{2.5cm}p{1.2cm}p{0.6cm}}
    \bottomrule
    \makecell{Task} & \makecell{Dataset} & \makecell{Input} & \makecell{Input\\Length} & \makecell{Output} & \makecell{Label\\Type} & Size \\
    \bottomrule
    \multirow{6}{*}{\makecell{Machine\\ Translation}} & MixMT2022\cite{srivastava-singh-2022-overview} & \makecell{English} & \makecell{16.85} & \makecell{English\\+Hindi;} & sentence&376\\
    & HinGE\cite{srivastava2021hinge} & \makecell{English}& \makecell{14.54} & \makecell{English\\+Hindi;} &sentence &2000\\
    & MT-AUG\cite{dhar-etal-2018-enabling} & \makecell{English} & \makecell{10.8} & \makecell{English\\+Hindi;} & sentence&610\\
      & PHINC\cite{srivastava-singh-2020-phinc} & \makecell{English} & \makecell{16.8} & \makecell{English\\+Hindi;} & sentence&400\\
    \hline
    \multirow{5}{*}{\makecell{Cross-Lingual\\ Transfer}}&  \multirow{5}{*}{\makecell{Multilingual\\ Controlled\\Generation\\~\cite{gupta2024multilingual}}} & \makecell{English} & \makecell{8.11}&\makecell{English\\+Bengali;} &sentence &549\\ 
    &  & \makecell{English} & \makecell{8.81}&\makecell{English\\+Spanish;} & sentence&350\\
    &  & \makecell{English} & \makecell{7.84}&\makecell{English\\+French;} &sentence &248\\
    \hline
    \multirow{4}{*}{\makecell{General\\ Classification}} & \makecell{SemEval-2020 \\ \cite{patwa2020semeval2020task9overview}} & \makecell{English\\+Hindi} & \makecell{26.12} & \makecell{English label} & \makecell{neg,\\ neutral,\\ pos} & 3000\\
    ~&\makecell{SentMix-3L\\ \cite{raihan2023sentmix}} & \makecell{English\\+Hindi} & \makecell{88.27} & \makecell{English label} & \makecell{neg,\\ neutral,\\ pos} &1007\\

  \bottomrule
\end{tabular}
\caption{Statistical information for code-mixing test sets used in the evaluation part.}
\label{tab:info_test_set}
\end{table*}

\subsection{Translation Metric Explanation}
\label{sec:appendix_metirc_explanation}
The introduction of these six well-studied metrics is below: (i) BLEU~\cite{papineni-etal-2002-bleu}, which measures the similarity between machine-translated text and human references based on n-gram precision and brevity penalties; (ii) \textit{chrF}~\cite{popovic2015chrf}, which calculates a character n-gram F-score based on the overlap between predicted and reference sentences; (iii) \textit{chrF++}~\cite{popovic2017chrf++}, which improves correlations with human assessment by adding word unigrams and bigrams to the standard chrF score; (iv) METEOR~\cite{banerjee-lavie-2005-meteor}, whose evaluation is based on the harmonic mean of unigram precision and recall; (v) COMET~\cite{rei2020comet}\footnote{We use reference-based evaluation model wmt22-comet-da to calculate the COMET score.}, which generates embeddings of the source, hypothesis, and reference sentences with a cross-lingual encoder~\cite{conneau2019unsupervised}, and predicts the score of the given translation; (vi) BERTScore~\cite{zhang2019bertscore}\footnote{xlm-roberta-large is used for BERTScore calculation}, which calculates token similarity using contextual embeddings from different variants of BERT.

\subsection{SFT Performance Degradation Analysis}
\label{sec:appendix_sft_performance_degradation}
Our results in Table~\ref{tab:performance_RLAIF_machine_translation} show that $\pi^{\mathrm{sft-2}}$ (i.e., SFT step applied on LlaMA-3.1-8B, the non-instruction tuned base version of our base LLM model) consistently outperforms $\pi^{\mathrm{sft-1}}$ (i.e., SFT step applied on LlaMA-3.1-8B-Instruct, the instruction tuned version of our base LLM model). This finding is consistent with prior work demonstrating that SFT on instruction-tuned models can induce performance degradation (as compared to SFT on base models) through catastrophic forgetting and distribution shift~\cite{ding2025improved,zhang2026best,wu2025shadow}.

\subsection{Statistical Analysis for Human Evaluation of Code-mixed Machine Translation in Figure~\ref{fig:performance_RLAIF_machine_translation_winrate_v3}}
\label{sec:appendix_stats}

For the human win rate comparison in Figure-2, we used a Two-Proportion Z-Test to calculate if the difference between our model and the base model (or SFT-2) is statistically significant. CHAI’s improvements over the base model was found to be statistically significant on both MT-AUG (p-value=0.0000001) and PHINC (p-value=0.0000001) datasets. Similarly, CHAI’s improvements over SFT-2 was found to be statistically significant on MT-AUG (p-value=0.001), but not significant on the PHINC dataset (p-value=0.36).

\subsection{Compare the Performance of RLAIF with DPO Pipeline}
\label{sec:appendix_comparision_dpo}
RLAIF pipeline and direct preference optimization (DPO) pipeline represent two prominent approaches for utilizing feedbacks to improve the performance of LLMs. By reformulating the objective function, DPO~\cite{rafailov2023direct} eliminates the need for an explicit reward model, and is therefore often considered a simplified and efficient alternative to RLHF.  We conduct an ablation study to compare the effectiveness of the RLAIF and DPO pipelines in improving translation performance under comparable training conditions (Training details could be checked at Appendix~\ref{sec:appendix_training_details_dpo}).

Table~\ref{tab:performance_compare_dpo} shows the translation performance in CHAI-LLM and $\pi^{dpo}$. The results based on MixMT-2022 indicate that translations generated by the CHAI-powered LLM achieve a win rate improvement of 40.08\% over the dpo model ($\pi^{\text{dpo}}$) according to LLM-based evaluators, and a 45.04\% improvement according to human evaluators. In terms of automatic metrics, CHAI-LLM surpasses $\pi^{\text{dpo}}$ with a 25.01\% increase in ChrF, a 24.28\% increase in ChrF++, and a 24.56\% improvement in COMET score.

As shown in the HinGE test set, the CHAI-powered LLM demonstrates substantial gains over the dpo model ($\pi^{\text{dpo}}$), achieving a 46.3\% higher win rate according to LLM-based evaluations and a 46.2\% higher win rate based on human judgments. Furthermore, across standard automatic metrics, CHAI-LLM consistently outperforms the baseline, with improvements of 81.57\% in ChrF, 71.02\% in ChrF++, and 69.04\% in COMET scores. In a word, Table~\ref{tab:performance_compare_dpo} suggests that the RLAIF pipeline is a more effective approach for enhancing the performance of LLMs in code-mixed translation scenarios.

\begin{table}[t]
    \centering
    \begin{tabular}{p{1.0cm}p{1.2cm}p{1.9cm}p{1.5cm}}
        \toprule
        Dataset & Evaluator & Results & \makecell{En -> \\Hinglish} \\
        
        \midrule
        \multirow{12}{*}{\makecell{MixMT\\2022}} 
        & \multirow{2}{*}{Gemini} & $\pi^{dpo}$ & 29.96\%  \\
& & CHAI-LLM & 70.04\%  \\

\cmidrule(lr){2-4}
& \multirow{2}{*}{Human} & $\pi^{dpo}$ & 27.48\%  \\
& & CHAI-LLM & 72.52\%  \\
\cmidrule(lr){2-4}
& \multirow{2}{*}{chrF} & $\pi^{dpo}$  &34.14   \\
& & CHAI-LLM  & 42.68  \\
\cmidrule(lr){2-4}
& \multirow{2}{*}{chrF++} & $\pi^{dpo}$ &30.84   \\
& & CHAI-LLM  & 38.33  \\
\cmidrule(lr){2-4}
& \multirow{2}{*}{COMET} & $\pi^{dpo}$  &0.57 \\
& & CHAI-LLM  & 0.71  \\

        \midrule
        \multirow{12}{*}{HinGE} 
        & \multirow{2}{*}{Gemini} & $\pi^{dpo}$ & 26.85\%  \\
& & CHAI-LLM & 73.15\%  \\

\cmidrule(lr){2-4}
& \multirow{2}{*}{Human} & $\pi^{dpo}$ & 26.90\%  \\
& & CHAI-LLM & 73.10\%  \\
\cmidrule(lr){2-4}
& \multirow{2}{*}{chrF} & $\pi^{dpo}$  &33.28   \\
& & CHAI-LLM  & 43.92  \\
\cmidrule(lr){2-4}
& \multirow{2}{*}{chrF++} & $\pi^{dpo}$ &29.82   \\
& & CHAI-LLM  & 39.56  \\
\cmidrule(lr){2-4}
& \multirow{2}{*}{COMET} & $\pi^{dpo}$  &0.55 \\
& & CHAI-LLM  &0.71  \\

        \bottomrule
    \end{tabular}
    \caption{Translation quality evaluation (based on LlaMA-3.1-8b-Instruct): CHAI-LLM vs DPO pipeline.}
    \label{tab:performance_compare_dpo}
\end{table}

\subsection{Training Details of RLAIF Procedure}
\label{sec:appendix_b}

\noindent\textbf{SFT stage.} SFT model is initialized from LlaMA-3.1-8b-Instruct.We use 1/3 of training dataset, with a learning rate of 1.0e-4, training for 3 epochs. The maximum input length is set up as 1024.\\

\noindent\textbf{Reward model training stage.} The reward model is initialized from LlaMA-3.1-8b-Instruct. The whole training data is used to form the chosen-rejected pairs with translated results collected from the open-source dataset of code-mixed machine translation tasks. We train 3 epochs with the learning rate of 1.0e-4, warm up ratio of 0.1, and maximum input length of 1024.

\noindent\textbf{RL fine-tuning stage.} We use the LlaMA-3.1-8b-Instruct as the initial policy. We reuse the input from the training data during the reward model training phase as queries. During RL fine-tuning, we sample from LLM with a temperature T=0.6 and nucleus sampling top\_p=0.9 and limit the maximum of generated length to 512. We train the model with a batch size of 16 and the learning rate of 1.0e-5 for 5 epochs. We set up the $\beta$=0.04 for the KL divergence loss (this coefficient value is obtained through one ablation study, selecting the value that yielded the best performance for the final model).\\

\subsection{Training Details of the SFT Baseline}
\label{sec:appendix_training_details_sft}
We start with LlaMA-3.1-8B (or LlaMA-3.1-8B-Instruct) to train the SFT baselines. The training data also comes from the same open source datasets: MixMT 2022 shared task and  ALL-CS dataset. In both datasets, each data point consists of an English source sentence and may include multiple corresponding Hinglish translations. For training purposes, we use the English sentence as the input and select the first corresponding Hinglish translation as the target label. We train 1 epochs with the learning rate of 1.0e-4, warm up ratio of 0.1, and maximum input length of 1024.

\subsection{Training Details of the DPO Pipeline}
\label{sec:appendix_training_details_dpo}
We start with LlaMA-3.1-8B to train the DPO model. Based on the preference label obtained from GPT-40, we transform the whole RLAIF training set into the DPO format training set. We train 3 epochs with the learning rate of 5.0e-6, warm up ratio of 0.1, $\beta$  of 0.1, DPO loss function of sigmoid, and maximum input length of 1024.

\subsection{Additional Cross-lingual Transfer Experiment based on Llama-3.1}
\label{sec:appendix_additional_Cross_lingual_Transfer}
In the En→CM (Fr–En) direction, as shown in Table~\ref{tab:corss_lingual_transfer_additional}, the $\pi^{\text{sft-1}}$ model achieves the highest BLEU score (16.93), chrF score (50.92), and chrF++ score (46.51).

There is evidence suggesting that both Spanish and Hindi have undergone substantial Arabic lexical borrowing, attributable to Moorish rule in the Iberian Peninsula and Mughal rule in the Indian subcontinent, respectively~\cite{borland2024arabic,khansir2014impact}, whereas Arabic has exerted considerably less influence on French. Estimates from non-academic sources indicate that Spanish and Hindi have incorporated approximately 4,000 and 5,500~\cite{kuczkiewicz2012perso} Arabic-derived words, respectively, compared to only approximately 700 such words in French. For instance, the Arabic word \textit{naran} ("naranj") corresponds to the Spanish \textit{naranja} and the Hindi \textit{nārangī}, all denoting the fruit "orange."

These lexical correspondences may give rise to shared sub-word vocabulary between Spanish and Hindi, which could account for the superior cross-lingual transfer performance of CHAI — trained on Hindi-English data — when evaluated on Spanish-English text relative to French-English text.

\begin{table*}[!ht]
  \centering
  \scalebox{0.72}{
  \begin{tabular}{ccccccc}
    \bottomrule
     & \multicolumn{6}{c}{\makecell{En$\rightarrow$CM of Fr and En\\ English$\rightarrow$Code-mixing of French and English}} \\

    \cmidrule(lr){2-7}
    \cmidrule(lr){2-7}
     & BLEU & ChrF & CHrF++ & METEOR & COMET &BERTscore\\
    \bottomrule
    
    $\pi^{\mathrm{base}}$ & 3.25  & 34.88 & 31.52 & 0.35 & 0.67 & 0.89  \\
    \hline
    $\pi^{\mathrm{sft-1}}$ & \textbf{16.93} & \textbf{50.92} & \textbf{46.51} & \textbf{0.41} & \textbf{0.84} & \textbf{0.92} \\
     \hline
    $\pi^{\mathrm{sft-2}}$ &1.45  & 31.90 & 27.02  & 0.31 & 0.63 & 0.87  \\
     \hline
    CHAI-LLM & 1.61 & 22.07  &  19.85 & 0.39 & 0.77 & 0.90\\

  \bottomrule
\end{tabular}
 }
\caption{Cross-lingual transfer result based on one additional code-mixed language pair.}
\label{tab:corss_lingual_transfer_additional}
\end{table*}

\subsection{Code-mixed Metrics for the Code-mixed Machine Translation Task}
\label{sec:appendix_additional_code_mixed_metrics}

To further substantiate our argument, we present three additional code-mixing evaluation metrics for the code-mixed machine translation task in Figure~\ref{tab:performance_RLAIF_machine_translation}: (i) Code-Mixing Index (CMI)~\cite{srivastava2021challenges}; (ii) M-Index~\cite{barnett2000lides}; and (iii) I-Index~\cite{guzman2017metrics}.

However, a key practical limitation is that computing these metrics requires word-level language identification (LID) labels for each sentence in the evaluation set. Neither the datasets used for evaluation in our paper (MT-AUG and PHINC) 
provide such (ground truth) word-level LID labels. Given these challenges, we utilize GPT-5-mini to generate the word level language labels for each sentence in our test set, and calculate these three metrics based on these LLM-generated word level labels. The results (based on LLama-3.1-8B and Qwen-2.5-7B) are shown in Table~\ref{tab:performance_RLAIF_machine_translation_code_mixed_metric_llama} and Table~\ref{tab:performance_RLAIF_machine_translation_code_mixed_metric_qwen}, respectively. In particular, these tables show that CHAI-LLM achieves 9.71\% (and 59.64\%) higher CMI on the MT-AUG dataset based on Llama-3.1 (and Qwen-2.5), respectively. Similar findings emerge on the PHINC dataset as well, where CHAI-LLM achieves 15.95\% better CMI score as compared to the next best performing baseline (on average). Results with the other two metrics (M-Index and I-Index) show similar trends. The results strengthen our argument that CHAI performs best across a wide variety of metrics.

\begin{table*}[!ht]
  \centering
  \scalebox{0.70}{
  \begin{tabular}{ccccccc}
    \toprule
     & \multicolumn{3}{c}{MT-AUG} & \multicolumn{3}{c}{PHINC}\\

    \cmidrule(lr){2-4}
    \cmidrule(lr){5-7}
     & CMI (Code-mixing Index) & M-Index & I-Index 
     & CMI (Code-mixing Index) & M-Index & I-Index \\
    \bottomrule
    
    $\pi^{\mathrm{base}}$ & 19.91 &  0.47 & 0.18 &
    20.96 & 0.49 & 0.22\\
    \hline
    $\pi^{\mathrm{sft-1}}$ & 22.07 & 0.53 & 0.31 &
    22.48 & \textbf{0.63} & 0.29\\
     \hline
    $\pi^{\mathrm{sft-2}}$ & 21.43 & 0.49 &  0.29& 
    21.86 & 0.61 & 0.27\\
     \hline
    CHAI-LLM & \textbf{23.19} & \textbf{0.55} & \textbf{0.33} & 
    \textbf{23.32} &  0.62 & \textbf{0.34}\\

  \bottomrule
\end{tabular}
 }
\caption{Measuring CHAI’s ability (on Llama-3.1) on code mixed translation ability based on code-mixed metrics.}
\label{tab:performance_RLAIF_machine_translation_code_mixed_metric_llama}
\end{table*}

\begin{table*}[!ht]
  \centering
  \scalebox{0.70}{
  \begin{tabular}{ccccccc}
    \toprule
     & \multicolumn{3}{c}{MT-AUG} & \multicolumn{3}{c}{PHINC}\\

    \cmidrule(lr){2-4}
    \cmidrule(lr){5-7}
     & CMI (Code-mixing Index) & M-Index & I-Index 
     & CMI (Code-mixing Index) & M-Index & I-Index \\
    \bottomrule
    
    $\pi^{\mathrm{base}}$ &  20.98 & 0.47  & 0.20 &
     23.41 & 0.53 & 0.22 \\
    \hline
    $\pi^{\mathrm{sft-1}}$ & 21.23 & 0.49 & 0.28 &
     24.18 & 0.55 & 0.26 \\
     \hline
    $\pi^{\mathrm{sft-2}}$ & 20.97 & 0.48 & 0.28 & 
     26.63 & 0.61 & 0.25\\
     \hline
    CHAI-LLM & \textbf{33.60} & \textbf{0.56} & \textbf{0.45} & 
    \textbf{34.13} & \textbf{0.67} & \textbf{0.45}\\

  \bottomrule
\end{tabular}
 }
\caption{Measuring CHAI’s ability (on Qwen-2.5) on code mixed translation ability based on code-mixed metrics.}
\label{tab:performance_RLAIF_machine_translation_code_mixed_metric_qwen}
\end{table*}

\subsection{Evaluation Result based on Qwen-2.5}
\label{sec:appendix_qwen}
In this section, we replicate the experiments described in the Experimental Evaluation section using Qwen-2.5.

\noindent \textbf{Impact of CHAI on Translation Quality.} We also train a CHAI-powered LLM based on Qwen-2.5 using the optimal hyperparameters identified earlier to evaluate its effectiveness for code-mixed translation. Table \ref{tab:performance_RLAIF_machine_translation_4} compares the performance of CHAI-LLM with three baseline models across six evaluation metrics on the MT-AUG and PHINC test sets. The results show that CHAI-LLM consistently outperforms all three baselines across all six metrics (e.g., achieving a BLEU score of 6.26 and a COMET score of 0.76 on MT-AUG, and a BLEU score of 17.26 and a COMET score of 0.70 on PHINC). Similar performance trends are observed for both Llama 3.1 and Qwen-2.5.

\noindent \textbf{Cross-lingual Transferability.} We examine whether translation preferences learned during CHAI’s RLAIF procedure enhance cross-lingual transfer. In particular, we investigate whether similar conclusions can be drawn when using Qwen-2.5 as the base model, compared to results obtained with the Llama model family. We evaluate the same three translation directions: (i) English $\rightarrow$ English + Bengali (En+Be);(ii) English $\rightarrow$ English + Spanish (En+Es); and (iii) English $\rightarrow$ English + French (En+Fr).

The evaluation results for these three translation directions are presented in Table~\ref{tab:corss_lingual_transfer_qwen}. Across six automatic evaluation metrics, CHAI-LLM demonstrates strong and consistent performance. In particular, CHAI-LLM (based on Qwen-2.5) achieves better results in the same two (of the three) translation directions. These findings are consistent with those observed for the Llama model family, further supporting that the RLAIF procedure improves CHAI-LLM’s cross-lingual transfer across multiple code-mixed language pairs.

\noindent\textbf{Cross-Task Ability. } We investigate whether applying RLAIF (with the Qwen model family) for code-mixed (CM) translation improves an LLM’s general capability to handle additional CM tasks. Table~\ref{tab:performance_sa_qwen} compares the accuracy and F1 scores of our CHAI-powered LLM (Qwen-2.5) with three baseline models on two CM sentiment analysis datasets. The results show a similar trend to those observed with Llama 3.1 (reported in Table~\ref{tab:performance_sa}), where CHAI-powered models consistently achieve stronger performance on the same test sets.

\begin{table}[t]
  \centering
  \begin{tabular}{p{2.0cm}p{1.05cm}p{1.2cm}p{1.2cm}}
    \bottomrule
    Dataset & LLM & Accuracy & F1\_score \\
    \bottomrule
    \multirow{4}{*}{SemEval-2020}&$\pi^{\mathrm{base}}$ & 56.50\% & 53.93\%\\
    & $\pi^{\mathrm{sft-1}}$ & 59.13\% & 58.07\%\\ 
    & $\pi^{\mathrm{sft-2}}$ & 56.6\% & 54.01\%\\
    & CHAI & \textbf{64.57\%} & \textbf{64.17\%}\\
    \hline
    \multirow{4}{*}{SentMix-3L}&$\pi^{\mathrm{base}}$ & 56.11\% & 49.16\%\\ 
    & $\pi^{\mathrm{sft-1}}$ & 58.49\% & 51.15\%\\
    & $\pi^{\mathrm{sft-2}}$ & 57.23\% & 50.20\%\\
    & CHAI & \textbf{67.03\%} & \textbf{64.11\%}\\

  \bottomrule
\end{tabular}
\caption{Qwen Performance of CHAI on sentiment analysis.}
\label{tab:performance_sa_qwen}
\end{table}

\begin{table*}[!ht]
  \centering
  \scalebox{0.72}{
  \begin{tabular}{ccccccccccccc}
    \bottomrule
     & \multicolumn{6}{c}{MT-AUG} & \multicolumn{6}{c}{PHINC}\\

    \cmidrule(lr){2-7}
    \cmidrule(lr){8-13}
     & BLEU & ChrF & CHrF++ & METEOR & COMET &BERTscore &BLEU  & ChrF & CHrF++ & METEOR & COMET &BERTscore\\
    \bottomrule
    
    $\pi^{\mathrm{base}}$ & 1.78    & 20.07 & 17.14 & 0.13 & 0.61 & 0.86 & 
    10.17  & 26.22 &  22.52 & 0.19 & 0.61 & 0.86\\
    \hline
    $\pi^{\mathrm{sft-1}}$ & 1.28 & 14.42 & 12.78 & 0.11 & 0.61 & 0.87 & 
    13.89 & 29.56 & 26.19 & 0.22 &  0.62 & 0.88 \\
     \hline
    $\pi^{\mathrm{sft-2}}$ &1.97  & 23.90 & 21.25  & 0.17 & 0.61 & 0.87& 
    10.73  &  32.66 & 29.27 & 0.28 & 0.64 & 0.88 \\
     \hline
    CHAI-LLM & \textbf{6.26}  & \textbf{33.89}  &  \textbf{29.52} & \textbf{0.28} & \textbf{0.76} &\textbf{0.89}& 
    \textbf{17.26}  & \textbf{38.09} & \textbf{34.95} & \textbf{0.34} & \textbf{0.70} & \textbf{0.90}\\

  \bottomrule
\end{tabular}
 }
\caption{Measuring CHAI (based on Qwen2.5)’s ability in improving code mixed translation ability.}
\label{tab:performance_RLAIF_machine_translation_4}
\end{table*}


\begin{table*}[!ht]
    \centering
    \scalebox{0.85}{
    \begin{tabular}{lllcccc}
        \toprule
        \multirow{3}{*}{\textbf{\makecell{Original Translation\\ Direction}}} & \multirow{2}{*}{\textbf{Evaluator}} & \multirow{2}{*}{\textbf{Results}} & \multicolumn{3}{c}{\textbf{Translation Direction}} \\
        \cmidrule(lr){4-6}
&& & \makecell{En$\rightarrow$CM of \\Be and En} & \makecell{En$\rightarrow$CM of \\Fr and En} &  \makecell{En$\rightarrow$CM of \\Es and En } \\
        \midrule
        \multirow{22}{*}{En$\rightarrow$Hinglish} 

& \multirow{4}{*}{BLEU} & $\pi^{base}$  & 2.27 & 10.96 & 16.42 \\
& & $\pi^{\mathrm{sft-1}}$ & \textbf{3.42} & \textbf{19.71} & 17.85 \\
& & $\pi^{\mathrm{sft-2}}$ & 2.32 & 8.43 & 7.52 \\
& & CHAI-LLM  & 3.15 & 15.75 & \textbf{18.93} \\

\cmidrule(lr){2-6}
& \multirow{4}{*}{chrF} & $\pi^{base}$  & 12.51 & 45.54 & 46.69 \\
& & $\pi^{\mathrm{sft-1}}$ & 27.76 & \textbf{51.76} & 46.20 \\
& & $\pi^{\mathrm{sft-2}}$ & 23.51 & 44.18 & 38.27 \\
& & CHAI-LLM  & \textbf{28.50} & 51.67 & \textbf{49.77} \\

\cmidrule(lr){2-6}
& \multirow{4}{*}{chrF++} & $\pi^{base}$ & 11.40 & 41.03 & 43.53  \\
& & $\pi^{\mathrm{sft-1}}$ & 24.49 & \textbf{47.92} & 43.03 \\
& & $\pi^{\mathrm{sft-2}}$ & 21.20 & 40.14 & 35.73 \\
& & CHAI-LLM  & \textbf{26.41} & 47.13 & \textbf{46.50} \\

\cmidrule(lr){2-6}
& \multirow{4}{*}{METEOR} & $\pi^{base}$  & 0.11 & 0.34 & 0.41 \\
& & $\pi^{\mathrm{sft-1}}$ & 0.21 & \textbf{0.44} & 0.40\\
& & $\pi^{\mathrm{sft-2}}$ & 0.23 & 0.37 & 0.39 \\
& & CHAI-LLM  & \textbf{0.27} & \textbf{0.44} & \textbf{0.45} \\

\cmidrule(lr){2-6}
& \multirow{4}{*}{COMET} & $\pi^{base}$  & 0.51 & 0.77 & 0.78 \\
& & $\pi^{\mathrm{sft-1}}$ & 0.58 & \textbf{0.85} & 0.80 \\
& & $\pi^{\mathrm{sft-2}}$ & 0.56 & 0.81 & 0.79 \\
& & CHAI-LLM  & \textbf{0.65} & 0.82 & \textbf{0.81} \\

\cmidrule(lr){2-6}
& \multirow{4}{*}{BERTscore} & $\pi^{base}$  & 0.84 & 0.90 & 0.91  \\
& & $\pi^{\mathrm{sft-1}}$ & \textbf{0.86} & \textbf{0.92} & \textbf{0.92} \\
& & $\pi^{\mathrm{sft-2}}$ & 0.85 & 0.91 & 0.91 \\
& & CHAI-LLM  & \textbf{0.86} & \textbf{0.92} & \textbf{0.92}  \\

        \bottomrule
    \end{tabular}
    }
    \caption{Cross-lingual transfer result (of Qwen2.5) based on different code-mixed language pairs.}
    \label{tab:corss_lingual_transfer_qwen}
\end{table*}

\subsection{Translation Quality On Synthetic Data}
\label{sec:appendix_translation_quality_on_synthetic_data}

We build synthetic data based on the source english sentences in the HinGE dataset. Table~\ref{tab:translation_quality_on_refined_translation} shows the translation

\begin{table*}[!ht]
  \centering
  \scalebox{0.72}{
  \begin{tabular}{ccccccc}
    \bottomrule
     & \multicolumn{6}{c}{\makecell{ English$\rightarrow$Code-mixing of English and Hindi}} \\

    \cmidrule(lr){2-7}
    \cmidrule(lr){2-7}
     & BLEU & ChrF & CHrF++ & METEOR & COMET &BERTscore\\
    \bottomrule
    
    \makecell{Initial translation (Llama-3.1-8B-Instruct)} & 4.08  & 24.89 & 22.69 & 0.23 & 0.68 & 0.88  \\
    \hline
   
    Final revised translation (GPT-5-mini) & 8.78 & 32.73  &  30.49 & 0.37 & 0.75 & 0.90\\

  \bottomrule
\end{tabular}
 }
\caption{Translation quality comparison between the initial translation and refined translation on HinGE dataset.}
\label{tab:translation_quality_on_refined_translation}
\end{table*}

\subsection{Continued Performance Evaluation from Synthetic Data}
\label{sec:appendix_additional_evlaution_syntheric_data}

\begin{figure}[t]
  \includegraphics[width=0.90\linewidth]{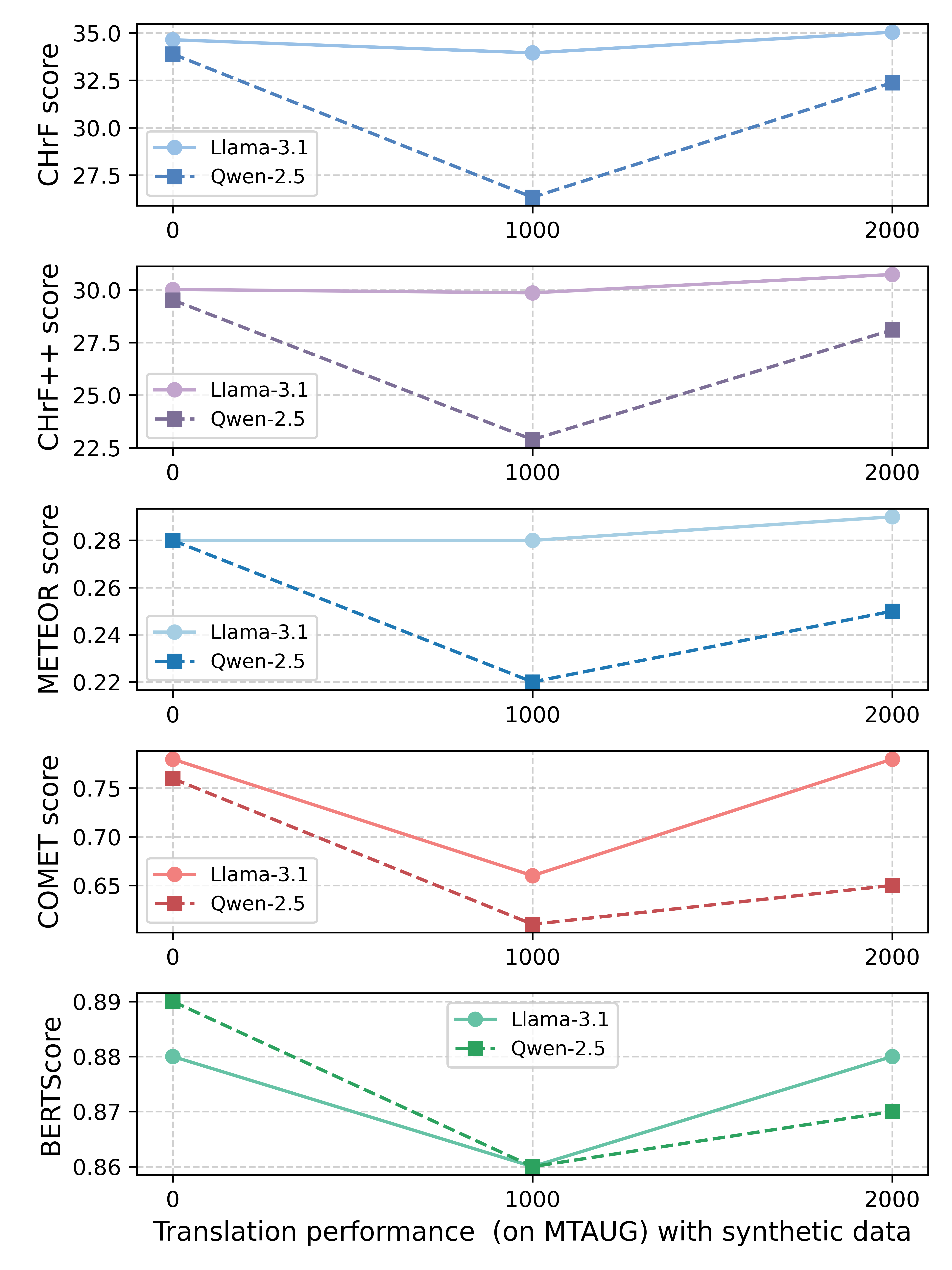}
  \caption {Evaluation based on MT-AUG test set comparing the performance gain vs. size of additional synthetyic data.}
  \label{fig:additonal_synthetic_data_on_MTAUG_5_metrics}
\end{figure}

\begin{figure}[t]
  \includegraphics[width=0.90\linewidth]{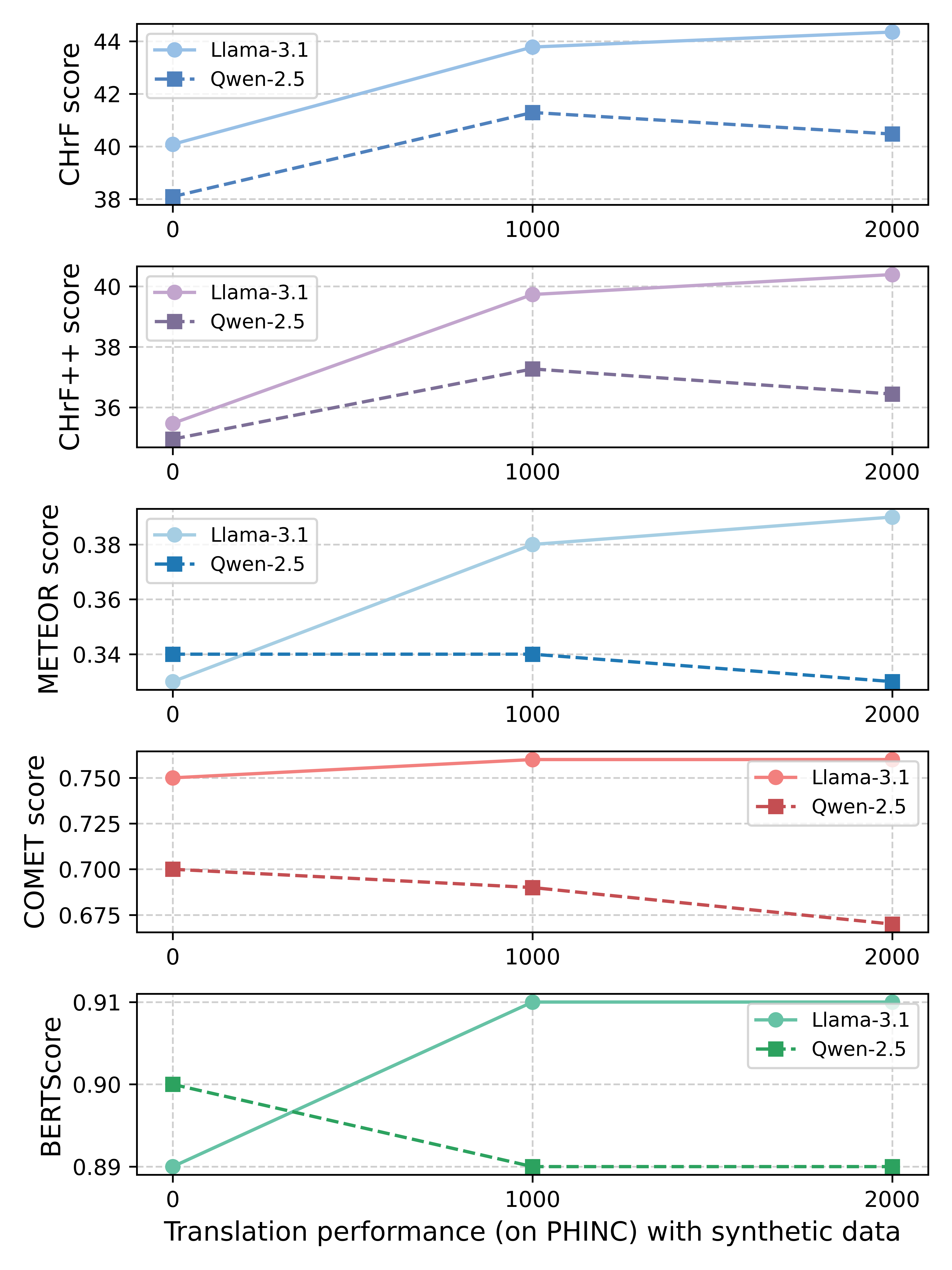}
  \caption {Evaluation based on PHINC test set comparing the performance gain vs. size of additional synthetyic data.}
  \label{fig:additonal_synthetic_data_on_PHINC_5_metrics}
\end{figure}

We compare the performance of the LLM trained on the original training set with that of the LLM trained on the original training set augmented with additional synthetic data. Figures~\ref{fig:additonal_synthetic_data_on_MTAUG_5_metrics} and \ref{fig:additonal_synthetic_data_on_PHINC_5_metrics} present additional automatic evaluation results on the MT-AUG and PHINC test sets, respectively.

\subsection{Analysis about the Alignment Score of the Evaluation Set}
 
We analyzed samples in which LLM and human choice differed.We analyze these samples and classify these disagreements into five categories (Examples for each of these samples are provided in Table \ref{tab:samples_machine_translation_disagreements}): 
\begin{enumerate}
  \item \textbf{Naturalness:} These translations differ mainly in their tone, choice of words, and naturalness while retaining the meaning of the sentence. The main distinguishing factor is whether the translation feels colloquial or not. For instance, some translations may appear awkward or disjointed due to poor integration of code-mixing.$\sim$68.6\% of samples were classified into this category.
  
  \item \textbf{Accuracy:} These translations differ in how they preserve the meaning and information of the original English sentence. A translation is considered inaccurate if it introduces new content, omits key information, alters the original meaning, or incorrectly uses code-switched terms.$\sim$10.2\% of the samples were classified into this category.
 
  \item \textbf{Code-switching correctness:} This category assesses whether the translation qualifies as a valid instance of code-switching. A sentence is considered to be code-mixed if (a) it is not entirely in Hindi or entirely in English, and (b) it does not include any language other than Hindi or English.$\sim$6.8\% of the samples were classified into this category.
 
  \item \textbf{Syntactic correctness:} This evaluates whether the translation adheres to grammatical and syntactic rules, including the proper integration of code-switched segments. Translations with clear syntax or grammar issues fall into this category.$\sim$2.4\% of the samples were classified into this category.
  
  \item \textbf{Low-quality data:} In these cases, both translation options were of poor quality and failed to adequately convey the original message. The lack of a clear, high-quality choice suggests that human preferences may have been arbitrary.$\sim$12.0\% of the samples were classified into this category.
 
\end{enumerate}

\begin{table*}[!ht]
  \centering
   \scalebox{0.85}{
  \begin{tabular}{p{2cm}|p{4cm}|p{8.7cm}}
    \bottomrule
     Category & Results & English -> Hinglish \\
    \bottomrule
     & Original Sentence (English) & The region was endemic for malaria, Kala-a-zar and leprosy.  \\
    Naturalness & Translation-0 & The kshetr was endemic for malaria, Kala-a-azar aur kushth rog.\\
     & Translation-1 & Yah region malaria, kala-a-azar and leprosy se prabhavit hai.  \\
     \cline{2-3}
     & Comments &  While both translations retain the meaning of the original sentence, participants preferred Translation-1 over Translation-0 because it reads more naturally and is closer to everyday spoken Hinglish. Translation-0 uses formal or Sanskritized terms like “kshetr” (region) and “kushth rog” (leprosy), which are less commonly used in colloquial speech, making it sound stilted and awkward. In contrast, Translation-1 blends Hindi and English more fluently and is easier to understand.\\
      \hline
     & Original Sentence (English) & We think doodling is something you do when you lose focus.  \\
    Accuracy & Translation-0 & We think doodling karne vaale ka dhyan bhang ho chuka he.\\
     & Translation-1 & Aesa mana jata he ki doodling is something you do when you lose focus.  \\
     \cline{2-3}
     & Comments &  Participants preferred Translation-1 over Translation-0 because it more accurately conveys the meaning of the original sentence. While Translation-0 paraphrases the intent, it subtly alters the meaning by implying that the person's attention has \textbf{already} been lost. In contrast, Translation-1 accurately captures the nuance of the original sentence, describing doodling as an activity that typically occurs \textbf{when} focus is lost.\\

      \hline
     & Original Sentence (English) & And what to do with all this?  \\
    Code-switching correctness & Translation-0 & And this all, inka kya karu? \\
     & Translation-1 & Aur yeh sab inka kya karu?  \\
     \cline{2-3}
     & Comments &  Participants preferred Translation-0 over Translation one since the former contains both Hindi and English words, in comparison to the latter, which contains only Hindi words. \\

      \hline

      & Original Sentence (English) & You won the TED Prize 2011.  \\
    Syntactic correctness & Translation-0 & tumhe TED Prize 2011 mil gaya hai.\\
     & Translation-1 & You won 2011 ka TED Prize.  \\
     \cline{2-3}
     & Comments &  Translation-1 was preferred for its natural and fluent code-mixed syntax. The use of "2011 ka TED Prize" aligns with common Hinglish patterns, while Translation-0 sounds more formal and slightly abrupt in structure.\\

      \hline

      & Original Sentence (English) & I eat very good food.  \\
    Low-quality data & Translation-0 & Mein bohot tasty food banata hu.\\
     & Translation-1 & Mein bohot aacha food banata hu.  \\
     \cline{2-3}
     & Comments &  Both translations misinterpret the original sentence, shifting the meaning from “I eat” to “I make” good food. Participants chose Translation-1 simply because it was slightly more coherent, but neither option accurately conveyed the intended message.\\
     
  \bottomrule
\end{tabular}
}
\caption{Representative examples of disagreements between human and LLM preferences, categorized by type of divergence}
\label{tab:samples_machine_translation_disagreements}
\end{table*}

\subsection{Adopted Version of the Constitutional AI Process}
\label{sec:appendix_Selected Principles}

In this section, we present the principles collected for the constitutional AI process under a code-switching scenario. Detailed examples are illustrated in Figure~\ref{fig:rules_under_constitutional_AI_processing}.

\begin{figure*}[t]
\tiny
\centering
\begin{tcolorbox}[colback=black!5!white,colframe=black!75!black,title=Principles under CM Scenario]

Rule 1 – Switch at syntactically permissible boundaries\\
\hspace*{0.5cm}• Only insert switches where both languages’ grammatical structures align (per Equivalence Constraint Theory).\\
\hspace*{0.5cm}• Ideal switch points: between clauses, noun–verb boundaries, or at prepositions and conjunctions.\\
\hspace*{0.5cm}• Avoid mid-word or intra-morpheme switches.\\

Rule 2 – Maintain the “Matrix Language Frame”\\
\hspace*{0.5cm}• Choose one language (e.g., Hindi) as the structural base (matrix language).\\
\hspace*{0.5cm}• Keep core morphology and word order in the matrix language while embedding content words from the embedded language (e.g., English nouns or adjectives).\\

Rule 3 – Prefer switching point based on judgement of POS tags\\
\hspace*{0.5cm}• Content words (NN, NNP, JJ) are the most natural switch points.\\
\hspace*{0.5cm}• Function words (VBZ, DT, IN, e.g., determiners, auxiliaries) should stay in the matrix language.\\
\hspace*{0.5cm}• Code-switches typically occur at phrase boundaries where meaning units are independent. The most common POS triggers for a language switch are nouns, verbs, and discourse markers. \\

Rule 4 – Retain culturally dominant or untranslatable English terms\\
\hspace*{0.5cm}• Maintain English for domain-specific, untranslatable, or culturally preferred words (e.g., internet, burger, app, charger, movie).\\
\hspace*{0.5cm}• Maintain Hindi for idiomatic expressions or sentiment-carrying words.\\

Rule 5 – Named Entity Knowledge\\
\hspace*{0.5cm}• Proper nouns (persons, places, organizations) rarely translated.\\
\hspace*{0.5cm}• For example, "Google," "New York," or "iPhone" should always be preserved in English; Entities such as doctor, WhatsApp, Amazon, Delhi are common English-retained tokens. \\

Rule 6 – Preserve semantic fidelity of the English source\\
\hspace*{0.5cm}• When regenerating the mixed sentence, prioritize semantic equivalence with the English source even if it requires reordering switches.\\
\hspace*{0.5cm}• Avoid literal translation of idioms—borrow the phrase in English if it conveys meaning more clearly (“heartbroken” → “dil toot gaya” not “heart toot gaya”).\\

Rule 7 – Respect topic-domain vocabulary\\
\hspace*{0.5cm}• In tech or academic domains, bias toward English switches (e.g., research project, mobile app).\\
\hspace*{0.5cm}• In casual/social domains, increase Hindi tokens for naturalness (Kya plan hai weekend ka?).\\

Rule 8 – Handle gender and agreement via matrix language rules\\
\hspace*{0.5cm}• Ensure Hindi verbs and adjectives agree in gender/number even if the noun is in English.\\

Rule 9 - Domain-Specific Lexical Knowledge\\
\hspace*{0.5cm}• Medical / Health Domain\\
        \hspace*{1cm}- Commonly retained English terms: doctor, nurse, hospital, blood pressure, test, diabetes, report, appointment.\\
        \hspace*{1cm}- Hindi usually wraps these in morphological markers, for example, Doctor ne check kiya vs Hospital mein admit hua.\\
        \hspace*{1cm}- Rule: Keep medical terminologies in English but ensure Hindi agreement (ne, mein, ko, etc.) aligns morphologically.\\
\hspace*{0.5cm}• Technology \& Internet Domain\\
        \hspace*{1cm}- Borrowed words: mobile, charger, Wi-Fi, internet, download, app, account, website.\\
        \hspace*{1cm}- Hindi words are often added for actions or contexts:
            o	App download karlo, Wi-Fi connect nahi ho raha.\\
        \hspace*{1cm}- Knowledge cue: Domain dictionary of untranslatable tech terms should remain unswitched.\\
\hspace*{0.5cm}• Education / Academia\\
        \hspace*{1cm}- Words like exam, class, result, assignment, research, topic typically stay in English.\\
        \hspace*{1cm}- Hindi words are used for connectors and verbs:
            o	Mera exam kal hai, Assignment submit kar diya kya?\\
        \hspace*{1cm}- Rule: Maintain English nouns, Hindi verbs and auxiliaries for naturalness.\\
\hspace*{0.5cm}• Finance / Commerce\\
        \hspace*{1cm}- Bank, account, payment, EMI, card, transaction often remain in English.\\
        \hspace*{1cm}- Code-mixing commonly occurs around numeric or financial terms. For example, Account balance check karna hai, Card decline ho gaya.\\

Rule 10 - Linguistic and Semantic Knowledge\\
\hspace*{0.5cm}• Morphological Integration Knowledge\\
        \hspace*{1cm}- English words adopt Hindi morphology\\
        \hspace*{1cm}- These patterns can be learned from bilingual dependency trees or surface patterns.\\

Rule 11 - Cultural \& Sociolinguistic Knowledge\\
\hspace*{0.5cm}• Borrowed Lexicon List\\
        \hspace*{1cm}- Maintain a curated list of English-origin words that have become part of colloquial Hindi (loanwords).
            o	Bus, train, ticket, police, college, cricket, TV, camera, phone.\\
        \hspace*{1cm}- Treat these as “unmarked switches” that should remain English in MT output.\\
\hspace*{0.5cm}• Honorific and Politeness Cues\\
        \hspace*{1cm}- Hindi uses politeness markers (aap, ji, kripya, zara).\\
        \hspace*{1cm}- When generating Hinglish, preserve these Hindi forms even around English insertions. For example, Aap please check kar lijiye.\\
        \hspace*{1cm}- Rule: Prioritize Hindi for politeness or formality markers regardless of surrounding English.\\
\hspace*{0.5cm}• Emotionally Loaded / Discourse Particles\\
        \hspace*{1cm}- Yaar, bhai, nahi, haan, please, sorry, okay are frequent switch points that convey tone.\\
        \hspace*{1cm}- Domain knowledge: preserve discourse particles in their most socially accepted language (often Hindi for emotion, English for politeness).

\end{tcolorbox}
\caption{Constitutional AI principles under the code-mixing scenario.}
    \label{fig:rules_under_constitutional_AI_processing}
\end{figure*}

\begin{table*}[]
    \centering
    \small
    \begin{tabular}{p{14.5cm}}
    \toprule

    \textbf{Prompt\_text:} You are a linguistics researcher specializing in code-switching. \\
    Write a set of clear annotation rules for improving natural code-switched sentences between two languages (e.g., English and Hindi).\\
    Requirements: \\
    1. Each rule should contain: \\
    \hspace*{0.5cm}- A short title starting with "Rule X – ..." \\
    \hspace*{0.5cm}- 2–3 bullet points explaining the rule. \\
    2. Avoid overly long explanations. Keep each bullet concise and practical. \\
    3. Write in the style of annotation guidelines for a dataset used in NLP research. \\\\
    Output format example: \\
    Rule 1 – [Rule title] \\
    \hspace*{0.5cm}• bullet point \\
    \hspace*{0.5cm}• bullet point \\
    Rule 2 – [Rule title] \\
    \hspace*{0.5cm}• bullet point \\
    \hspace*{0.5cm}• bullet point\\

    \bottomrule
    \end{tabular}
    \caption{Prompt for principles creation in CM constitution AI process.}
    \label{tab:prompt_for_ai_institution}
\end{table*}

\subsection{Commercial Model Usage}
\label{sec:Comercial_Model_Usage}
In this paper, we have utilized three commercial models: (i) gpt-4o model used for preference annotation refers "gpt-4o-2024-11-20"; (ii) Gemini model used for LLM evlauator points to "Gemini-2.5-flash-preview-09-2025"; (3) GPT-5-mini used in self-critique procedure refers "gpt-5-mini-2025-08-07".

\subsection{Human Evaluation Rules}
\label{sec:appendix_human_evaluation}
We set up human evaluation rules from four aspects: (i) accuracy; (ii) naturalness; (iii) syntactic correctness; (iv) code-switching Correctness. Details of each aspect are mentioned below.

\noindent\textbf{Accuracy.} It evaluates how effectively the translated sentence retains the meaning and information of the original sentence, while ensuring the correct usage of code-switched terms. For example, does the translation faithfully reflect the content of the original meaning? Is the key information missing, altered, or repeated in translated sentences? Does the translation introduce new information that is not covered in the original sentences?

\noindent\textbf{Naturalness.} It assesses how natural and easy to understand the translated sentence is. For example, is the new translation elegant? Does the translated sentence seem difficult to understand, awkward, or contain unnatural phrasing?

\noindent\textbf{Syntactic correctness.} It considers grammar, syntax, and the seamless integration of code-switching in translated sentences. Are there any grammar or syntax issues in translations? Does code-mixing disrupt the flow of the sentence? Is it somewhat smooth but not perfectly integrated? Or is it smooth and seamless?

\noindent\textbf{Code-switching Correctness.} It evaluates whether the given sentence is a correct instance of code-switching (CS). Specifically, we define a sentence as a correct CS sentence if it meets the following constraints: (a) it is not entirely in Hindi or English, and (b) no language other than Hindi or English is used.

\subsection{Prompts for Preference Labeling}
\label{sec:appendix_prompts_for_preference_labeling}
See different prompt strategies at Table~\ref{tab:prompt_preference_annotation_zero_shot_basic}, Table~\ref{tab:prompt_preference_annotation_zero_shot_detailed}, Table~\ref{tab:prompt_preference_annotation_cot_basic}, and Table~\ref{tab:prompt_preference_annotation_1_shot}.

\begin{table*}[]
    \centering
    \small
    \begin{tabular}{p{14.5cm}}
    \toprule

    \textbf{Prompt\_text:} You are a fluent Hinglish speaker. Fluent Hinglish speakers are able to switch between Hindi and English in the same sentence effortlessly while having a conversation.\\
You have an English sentence for which you’d like to choose the best Hinglish translation. \\
The English sentence is: \{original\_sent\};\\
Translated-sentence-0 is: \{first\_translation\};\\
Translated-sentence-1 is: \{second\_translation\};\\\\

Choose a translated statement that best aligns with how a fluent Hinglish speaker talks.
The format of the output should be as follows: “My preference is:”,followed by the number 0 or 1 (which signifies the corresponding translated sentence) based on your preference.\\

    \bottomrule
    \end{tabular}
    \caption{Basic zero-shot prompt for preference labeling on code-mixed texts.}
    \label{tab:prompt_preference_annotation_zero_shot_basic}
\end{table*}

\begin{table*}[]
    \centering
    \small
    \begin{tabular}{p{14.5cm}}
    \toprule

    \textbf{Prompt\_text:} A good code-mixed translation seamlessly blends elements of two or more languages while maintaining the original meaning and context. 
It ensures clarity and fluency in both languages, allowing the message to be easily understood by speakers of all involved languages.\\
Below we define four evaluation axes for code-mixed translation quality: accuracy, naturalness, syntactic correctness, and Code-switching Correctness.

1.Accuracy: It evaluates how effectively the translated sentence retains the meaning and information of the original sentence, while ensuring the correct usage of code-switched terms. For example, does the translation faithfully reflect the content of the original meaning? Is the key information missing, alternated or repeated in translated sentences? Does the translation introduce the new information which are not covered in original sentences?

2.Naturalness: It assesses how natural and easy to understand the translated sentence is. For example, is the new translation elegant? Does the translated sentence seem difficult to understand, awkward, or contain unnatural phrasing?

3.Syntactic correctness: It considers grammar, syntax, and the seamless integration of code-switching in translated sentences. Are there any grammar or syntax issues in translation? Does code-mixing disrupt the flow of the sentence? Is it somewhat smooth but not perfectly integrated? Or is it smooth and seamless?

4.Code-switching Correctness: It evaluates whether the given sentence is a correct instance of code-switching (CS). Specifically, we define a sentence as a correct CS sentence if it meets the following constraints: (a) it is not entirely in Hindi or English, and (b) no language other than Hindi or English is used.\\\\\\
    
    You are a fluent Hinglish speaker. Fluent Hinglish speakers are able to switch between Hindi and English in the same sentence effortlessly while having a conversation.\\
You have an English sentence for which you’d like to choose the best Hinglish translation. \\
The English sentence is: \{original\_sent\};\\
Translated-sentence-0 is: \{first\_translation\};\\
Translated-sentence-1 is: \{second\_translation\};\\\\

Choose a translated statement that best aligns with how a fluent Hinglish speaker talks.
The format of the output should be as follows: “My preference is:”,followed by the number 0 or 1 (which signifies the corresponding translated sentence) based on your preference.\\

    \bottomrule
    \end{tabular}
    \caption{rule-augmented zero-shot prompt for preference labeling on code-mixed texts.}
    \label{tab:prompt_preference_annotation_zero_shot_detailed}
\end{table*}

\begin{table*}[]
    \centering
    \small
    \begin{tabular}{p{14.5cm}}
    \toprule

    \textbf{Prompt-1 (output\_rationale):} You are a fluent Hinglish speaker. Fluent Hinglish speakers are able to switch between Hindi and English in the same sentence effortlessly while having a conversation.\\
You have an English sentence and two of its possible Hinglish translation. \\
Explain the reason that which translation is better.\\
The format of the output should be as follows: “Rationale:”,followed by the reasons in one paragraph.\\\\
The English sentence is: \{original\_sent\};\\
Translated-sentence-0 is: \{first\_translation\};\\
Translated-sentence-1 is: \{second\_translation\};\\

\hline
\textbf{Prompt-2 (output\_preference):} You are a fluent Hinglish speaker. Fluent Hinglish speakers are able to switch between Hindi and English in the same sentence effortlessly while having a conversation.\\
You have an English sentence, two of its possible Hinglish translation, and corresponding rationale. \\
Choose a translated statement that best aligns with how a fluent Hinglish speaker talks.\\
The format of the output should be as follows: “My preference is:”,followed by the number 0 or 1 (which signifies the corresponding translated sentence) based on your preference.\\\\

The English sentence is: \{original\_sent\};\\
Translated-sentence-0 is: \{first\_translation\};\\
Translated-sentence-1 is: \{second\_translation\};\\
Rationale: \{rationale\}\\
    \bottomrule
    \end{tabular}
    \caption{Basic zero-shot chain-of-thought prompt~\cite{lee2024rlaifvsrlhfscaling} for preference labeling on code-mixed texts, where we first generate the rational based on prompt-1 and then concatenate it with prompt-2 to generate the final preference label.}
    \label{tab:prompt_preference_annotation_cot_basic}
\end{table*}

\begin{table*}[]
    \centering
    \small
    \begin{tabular}{p{14.5cm}}
    \toprule

    \textbf{Prompt:} You are a fluent Hinglish speaker. Fluent Hinglish speakers are able to switch between Hindi and English in the same sentence effortlessly while having a conversation.\\
You have an English sentence for which you’d like to choose the best Hinglish translation. \\
Choose a translated statement that best aligns with how a fluent Hinglish speaker talks.\\
You could only output 0 (if you prefers Translated-sentence-0) or output 1 (if you prefers Translated-sentence-1)\\\\

»»»» Example »»»»\\
The English sentence is: <original\_sent for example-1>;\\
Translated-sentence-0 is: <first\_translationfor example-1>;\\
Translated-sentence-1 is: <second\_translation for example-1>;\\
My preference is: <label for example-1>\\\\

»»»» Follow the instructions and the example(s) above »»»»\\
The English sentence is: \{original\_sent\};\\
Translated-sentence-0 is: \{first\_translation\};\\
Translated-sentence-1 is: \{second\_translation\};\\
My preference is:\\

    \bottomrule
    \end{tabular}
    \caption{Basic 1-shot prompt for preference labeling on code-mixed texts.}
    \label{tab:prompt_preference_annotation_1_shot}
\end{table*}

\subsection{Prompt for LLM-based Evaluation}
\label{sec:appendix_e}
See LLM evaluation prompt at Table~\ref{tab:prompt_llm_as_judge}.

\begin{table*}[]
    \centering
    \small
    \begin{tabular}{p{14.5cm}}
    \toprule
    \textbf{System\_role:} You are a translation expert in \{source\_language\}, \{target\_language\}, code-mixing of \{source\_language\} and \{target\_language\}. I need your help in impartially judging the quality of two translations.\\

    \textbf{Prompt\_text:} Below we define four evaluation axes for code-mixed translation quality: accuracy, naturalness, syntactic correctness, and Code-switching Correctness.\\
1.Accuracy: It evaluates how effectively the translated sentence retains the meaning and information of the original sentence, while ensuring the correct usage of code-switched terms. For example, does the translation faithfully reflect the content of the original meaning? Is the key information missing, alternated or repeated in translated sentences? Does the translation introduce the new information which are not covered in original sentences?\\
2.Naturalness: It assesses how natural and easy to understand the translated sentence is. For example, is the new translation elegant? Does the translated sentence seem difficult to understand, awkward, or contain unnatural phrasing?\\
3.Syntactic correctness: It considers grammar, syntax, and the seamless integration of code-switching in translated sentences. Are there any grammar or syntax issues in translation? Does code-mixing disrupt the flow of the sentence? Is it somewhat smooth but not perfectly integrated? Or is it smooth and seamless?\\
4.Code-switching Correctness: It evaluates whether the given sentence is a correct instance of code-switching (CS). Specifically, we define a sentence as a correct CS sentence if it meets the following constraints: (a) it is not entirely in Hindi or English, and (b) no language other than Hindi or English is used.\\\\

Next, I will provide you with the original text under the <Original> tag, first translation under the <Translation\_1>, and second translation under the <Translation\_2>. \\
Please let me know which one is better according to these criteria. Please give your judgment directly (output "Translation\_1" or "Translation\_2" only) and do not output additional explanations.\\
<Original>\\
\{original\_sent\}\\
</Original>\\\\

<Translation\_1>\\
\{first\_translation\}\\
</Translation\_1>\\\\

<Translation\_2>\\
\{second\_translation\}\\
</Translation\_2>\\
    \bottomrule
    \end{tabular}
    \caption{Prompt for LLM-based evaluation.}
    \label{tab:prompt_llm_as_judge}
\end{table*}

\subsection{Human Evaluation \& Annotator Agreement}
\label{sec:appendix_f}

Expert annotators fluent in both English and Hindi were recruited from India to conduct the human evaluation. All annotators possessed the linguistic competence necessary to make rigorous assessments of fluency and accuracy in code-switching contexts.

Inter-annotator agreement is measured using Fleiss' kappa~\cite{fleiss1971measuring}, a statistical metric designed to assess the degree of consensus among multiple raters. Fleiss' kappa values range from 0 to 1, where 0 denotes agreement equivalent to chance and 1 represents perfect agreement. Our human evaluation yielded a Fleiss' kappa score of 0.43. We consider this a reasonable level of agreement given that annotating code-mixing is a significantly more complex task than monolingual annotation (because there is no formal grammar that governs code-mixed language, and hence it is harder to define a correct annotation for code-mixed translation).

\end{document}